\documentclass[11pt]{article}

\usepackage{fullpage}
\usepackage[round,comma]{natbib}
\usepackage{graphicx}
\usepackage{subfigure}
\usepackage{algorithm}
\usepackage{algorithmic}
\usepackage{amsmath}
\usepackage{amsfonts}
\usepackage{amssymb}

\usepackage{url}
\usepackage{color}
\definecolor{greytext}{gray}{0.5}

\usepackage{amsmath,amsthm,amssymb}

\usepackage{color}
\usepackage[all]{xy}
\usepackage{verbatim}
\usepackage{cancel}
\definecolor{Blue}{rgb}{0.9,0.3,0.3}

\newcommand{\squishlist}{
   \begin{list}{$\bullet$}
    { \setlength{\itemsep}{0pt}      \setlength{\parsep}{3pt}
      \setlength{\topsep}{3pt}       \setlength{\partopsep}{0pt}
      \setlength{\leftmargin}{1.5em} \setlength{\labelwidth}{1em}
      \setlength{\labelsep}{0.5em} } }

\newcommand{\squishlisttwo}{
   \begin{list}{$\bullet$}
    { \setlength{\itemsep}{0pt}    \setlength{\parsep}{0pt}
      \setlength{\topsep}{0pt}     \setlength{\partopsep}{0pt}
      \setlength{\leftmargin}{2em} \setlength{\labelwidth}{1.5em}
      \setlength{\labelsep}{0.5em} } }

\newcommand{\squishend}{
    \end{list}  }

\newcommand{\Ber}{\mathrm{Ber}}

\newcommand{\myvec}[1]{\mathbf{#1}}
\newcommand{\myvecsym}[1]{\boldsymbol{#1}}
\newcommand{\ind}[1]{\mathbb{I}(#1)}

\newcommand{\vtheta}{\myvecsym{\theta}}

\newcommand{\sigmoid}{\mathrm{sigm}}

\newcommand{\vh}{\myvec{h}}

\newcommand{\vw}{\myvec{w}}

\newcommand{\vx}{\myvec{x}}

\newcommand{\vG}{\myvec{G}}

\newcommand{\vI}{\myvec{I}}

\newcommand{\vK}{\myvec{K}}

\newcommand{\vP}{\myvec{P}}
\newcommand{\vQ}{\myvec{Q}}

\newcommand{\be}{\begin{equation}}
\newcommand{\ee}{\end{equation}}
\newcommand{\bea}{\begin{eqnarray}}
\newcommand{\eea}{\end{eqnarray}}
\newcommand{\beaa}{\begin{eqnarray*}}
\newcommand{\eeaa}{\end{eqnarray*}}

\DeclareMathAlphabet{\mathpzc}{OT1}{pzc}{m}{n}

\newcommand{\fib}[3]{%
 \xy(0,5) \xymatrix@=+10pt{%
 #1 \ar[d]_{#2} \\
 #3             } \endxy }

\newcommand{\mapdef}[5]{%
\xy(0,4) \xymatrix@R=+0pt{%
 \negthickspace\negthickspace\negthickspace    #1 : #2    \ar[r]        &  #3  \\
 \quad #4                                                 \ar@{|->}[r]  &  #5  } \endxy }

\newcommand{\mapdefDisp}[5]{%
\xymatrix@R=+0pt{%
 \negthickspace\negthickspace\negthickspace    #1 : #2    \ar[r]        &  #3  \\
 \quad #4                                                 \ar@{|->}[r]  &  #5  } }

\def\abovestrut#1{\rule[0in]{0in}{#1}\ignorespaces}
\def\belowstrut#1{\rule[-#1]{0in}{#1}\ignorespaces}

\def\abovespace{\abovestrut{0.20in}}

\def\belowspace{\belowstrut{0.10in}}

\title{A Machine Learning Perspective on Predictive Coding with PAQ}

\author{Byron Knoll \& Nando de Freitas\\
  University of British Columbia \\
  Vancouver, Canada \\
  \texttt{\small \{knoll,nando\}@cs.ubc.ca}
}

\date{\today}

\begin{document}

\newcommand{\Exh}{E_{\vtheta}(\vx,\vh)}
\newcommand{\Fx}{F_{\vtheta}(\vx)}

\newcommand{\sumi}{\sum_{i=1}^{n_x}}
\newcommand{\sumj}{\sum_{j=1}^{n_{h}^m}}
\newcommand{\sumk}{\sum_{k=1}^{n_{h}^c}}
\newcommand{\sumf}{\sum_{f=1}^{n_f}}
\newcommand{\sumip}{\sum_{i'=1}^{n_x}}

\maketitle

\begin{abstract}
PAQ8 is an open source lossless data compression algorithm that currently achieves the best compression rates on many benchmarks. This report presents a detailed description of PAQ8 from a statistical machine learning perspective. It shows that it is possible to understand some of the modules of PAQ8 and use this understanding to improve the method. However, intuitive statistical explanations of the behavior of other modules remain elusive. We hope the description in this report will be a starting point for discussions that will increase our understanding, lead to improvements to PAQ8, and facilitate a transfer of knowledge from PAQ8 to other machine learning methods, such a recurrent neural networks and stochastic memoizers.
Finally, the report presents a broad range of new applications of PAQ to machine learning tasks including language modeling and adaptive text prediction, adaptive game playing, classification, and compression using features from the field of deep learning.
\end{abstract}



\section{Introduction}
\label{ch:Introduction}

Detecting temporal patterns and predicting into the future is a fundamental problem in machine learning. It has gained great interest recently in the areas of nonparametric Bayesian statistics \citep{Wood09} and deep learning \citep{Sutskever11}, with applications to several domains including language modeling and unsupervised learning of audio and video sequences. Some researchers have argued that sequence prediction is key to understanding human intelligence \citep{Hawkins2005}.

The close connections between sequence prediction and data compression are perhaps underappreciated within the machine learning community. The goal of this report is to describe a state-of-the-art compression method called \textit{PAQ8} \citep{Mahoney05} from the perspective of machine learning. We show both how PAQ8 makes use of several simple, well known machine learning models and algorithms, and how it can be improved by exchanging these components for more sophisticated models and algorithms. 

PAQ is a family of open-source compression algorithms closely related to the better known Prediction by Partial Matching (PPM) algorithm \citep{Cleary84}. PPM-based data compression methods dominated many of the compression benchmarks (in terms of compression ratio) in the 1990s, but have since been eclipsed by PAQ-based methods. Compression algorithms typically need to make a trade-off between compression ratio, speed, and memory usage. PAQ8 is a version of PAQ which achieves record breaking compression ratios at the expense of increased time and memory usage. For example, all of the winning submissions in the Hutter Prize \citep{Hutter06}, a contest to losslessly compress the first 100 MB ($10^{8}$ bytes) of Wikipedia, have been specialized versions of PAQ8. Dozens of variations on the basic PAQ8 method can be found on the web:  \emph{http://cs.fit.edu/$\sim$mmahoney/compression/paq.html}. As stated on the Hutter Prize website, ``This compression contest is motivated by the fact that being able to compress well is closely related to acting intelligently, thus reducing the slippery concept of intelligence to hard file size numbers.''

\begin{table*}[t]
\caption{Comparison of cross entropy rates of several compression algorithms on the Calgary corpus files. The cross entropy rate metric is defined in Section~\ref{metrics}.}\
\label{tab:a}
\begin{center}
\begin{small}
\begin{sc}
{\begin{tabular}{ lcccccc}
\hline
\abovespace\belowspace		
File & PPM-test & PPM*C & 1PF & UKN & cPPMII-64 & PAQ8L\\
\hline
\abovespace
bib	&1.80	&1.91	&1.73 & 1.72 &1.68 & 1.50\\
book1	&2.21	&2.40	&2.17 & 2.20 &2.14 & 2.01\\
book2	&1.88	&2.02	&1.83 & 1.84 &1.78 & 1.60\\
geo	&4.65	&4.83	&4.40 & 4.40 &4.16 & 3.43\\
news	&2.28	&2.42	&2.20 & 2.20 &2.14 & 1.91\\
obj1	&3.87	&4.00	&3.64 & 3.65 &3.50 & 2.77\\
obj2	&2.37	&2.43	&2.21 & 2.19 &2.11 & 1.45\\
paper1	&2.26	&2.37	&2.21 & 2.20 &2.14 & 1.97\\
paper2	&2.22	&2.36	&2.18 & 2.18 &2.12 & 1.99\\
pic	&0.82	&0.85	&0.77 & 0.82 &0.70 & 0.35\\
progc	&2.30	&2.40	&2.23 & 2.21 &2.16 & 1.92\\
progl	&1.57	&1.67	&1.44 & 1.43 &1.39 & 1.18\\
progp	&1.61	&1.62	&1.44 & 1.42 &1.39 & 1.15\\
trans	&1.35	&1.45	&1.21 & 1.20 &1.17 & 0.99\\
\belowspace
{\bf Average}	&{\bf 2.23}	&{\bf 2.34}	&{\bf 2.12} & {\bf 2.12} &{\bf 2.04} & {\bf 1.73}\\
\hline
\end{tabular}}
\end{sc}
\end{small}
\end{center}
\vskip -0.1in
\end{table*}

\begin{table*}[t]
\caption{File size and description of Calgary corpus files.}
\label{calgary}
\begin{center}
\begin{small}
\begin{sc}
\addtolength{\tabcolsep}{-2pt}
{\begin{tabular}{ lll}
\hline
\abovespace\belowspace		
File & bytes & Description\\
\hline
\abovespace
bib	&111,261 & {\scriptsize ASCII text in UNIX ``refer'' format - 725 bibliographic references.}\\
book1	&768,771 & {\scriptsize unformatted ASCII text - ``Far from the Madding Crowd''}\\
book2	&610,856 & {\scriptsize ASCII text in UNIX ``troff'' format - ``Principles of Computer Speech''}\\
geo	&102,400 & {\scriptsize 32 bit numbers in IBM floating point format - seismic data.}\\
news	&377,109 & {\scriptsize ASCII text - USENET batch file on a variety of topics.}\\
obj1	&21,504 & {\scriptsize VAX executable program - compilation of PROGP.}\\
obj2	&246,814 & {\scriptsize Macintosh executable program - ``Knowledge Support System''.}\\
paper1	&53,161 & {\scriptsize ``troff'' format - Arithmetic Coding for Data Compression.}\\
paper2	&82,199 & {\scriptsize ``troff'' format - Computer (in)security.}\\
pic	&513,216 & {\scriptsize 1728 x 2376 bitmap image (MSB first).}\\
progc	&39,611 & {\scriptsize Source code in C - UNIX compress v4.0.}\\
progl	&71,646 & {\scriptsize Source code in Lisp - system software.}\\
progp	&49,379 & {\scriptsize Source code in Pascal - program to evaluate PPM compression.}\\
\belowspace
trans	&93,695 & {\scriptsize ASCII and control characters - transcript of a terminal session.}\\
\hline
\end{tabular}}
\end{sc}
\end{small}
\end{center}
\vskip -0.1in
\end{table*}

The stochastic sequence memoizer \citep{Wood2010} is a language modeling technique recently developed in the field of Bayesian nonparametrics. Table~\ref{tab:a} shows a comparison of several compression algorithms on the Calgary corpus \citep{Bell1990}, a widely-used compression benchmark. A summary of the Calgary corpus files appears in Table~\ref{calgary}. {\tt PPM-test} is our own PPM implementation used for testing different compression techniques. {\tt PPM*C} is a PPM implementation that was state of the art in 1995 \citep{Cleary1995}. {\tt 1PF} and {\tt UKN} are implementations of the stochastic sequence memoizer \citep{Wood2010}. {\tt cPPMII-64} \citep{Shkarin2002} is currently among the best PPM implementations. {\tt paq8l} outperforms all of these compression algorithms by what is considered to be be a very large margin in this benchmark.

Despite the huge success of PAQ8, it is rarely mentioned or compared against in machine learning papers. There are reasons for this. A core difficulty is the lack of scientific publications on the inner-workings of PAQ8. To the best of our knowledge, there exist only incomplete high-level descriptions of PAQ1 through 6 \citep{Mahoney05} and PAQ8 \citep{Mahoney11}. The C++ source code, although available, is very close to machine language (due to optimizations) and the underlying algorithms are difficult to extract. Many of the architectural details of PAQ8 in this report were understood by examining the source code and are presented here for the first time.

\subsection{Contributions}

We provide a detailed explanation of how PAQ8 works. We believe this contribution will be of great value to the machine learning community. An understanding of PAQ8 could lead to the design of better algorithms. As stated in \citep{Mahoney11}, PAQ was inspired by research in neural networks: ``Schmidhuber and Heil (1996) developed an experimental neural network data compressor. It used a 3 layer network trained by back propagation to predict characters from an 80 character alphabet in text. It used separate training and prediction phases. Compressing 10 KB of text required several days of computation on an HP 700 workstation.'' In 2000, Mahoney made several improvements that made neural network compression practical. His new algorithm ran $10^5$ times faster. PAQ8 uses techniques (\emph{e.g.} dynamic ensembles) that could lead to advances in machine learning. 

As a second contribution, we demonstrate that an understanding of PAQ8 enables us to deploy machine learning techniques to achieve better compression rates. Specifically we show that a second order adaptation scheme, the extended Kalman filter (EKF), results in improvements over PAQ8's first order adaptation scheme.

A third contribution is to present several novel applications of PAQ8. First, we demonstrate how PAQ8 can be applied to adaptive text prediction and game playing. Both of these tasks have been tackled before using other compression algorithms. Second, we show for the first time that PAQ8 can be adapted for classification. Previous works have explored using compression algorithms, such as RAR and ZIP, for classification \citep{Marton05}. We show that our proposed classifier, PAQclass, can outperform these techniques on a text classification task. We also show that PAQclass achieves near state-of-the-art results on a shape recognition task. Finally, we develop a lossy image compression algorithm by combining PAQ8 with recently developed unsupervised feature learning techniques.

\subsection{Organization of this Report}

In Section~\ref{ch:lossless} we provide general background information about the problem of lossless data compression, including a description of arithmetic coding and PPM. In Section~\ref{ch:PAQ8} we present a detailed explanation of how PAQ8 works. This section also includes a description of our improvement to the compression rate of PAQ8 using EKF. We present several novel applications of PAQ8 in Section~\ref{ch:Applications}. Section~\ref{ch:Conclusion} contains our conclusions and possible future work. Appendix ~\ref{ch:a} contains information on how to access demonstration programs we created using PAQ8.


\section{Lossless Data Compression}
\label{ch:lossless}

An arbitrary data file can be considered as a sequence of characters in an alphabet. The characters
could be bits, bytes, or some other set of characters (such as ASCII or Unicode characters). Lossless data
compression usually involves two stages. The first is creating a probability distribution for the
prediction of every character in a sequence given the previous characters. The second is to encode these probability distributions into a file using
a coding scheme such as arithmetic coding \citep{Rissanen79} or Huffman coding \citep{Huffman52}. Arithmetic coding is usually preferable to Huffman coding because arithmetic coders can produce near-optimal encodings for any set of symbols and probabilities (which is not true of Huffman coders).  Since the problem of encoding predicted probability distributions to a file has been solved, the performance difference between compression algorithms is due to the way that they assign probabilities to these predictions. In the next section we give an overview of how arithmetic coding works.

\subsection{Arithmetic Coding}

Arithmetic coders perform two operations: encoding and decoding. The encoder takes as input a sequence of characters and a sequence of predicted probability distributions. It outputs a compressed representation of the sequence of characters. The decoder takes as input the compressed representation and a sequence of predicted probability distributions. It outputs the original sequence of characters (exactly equivalent to the encoder's input). The sequence of predicted probability distributions needs to be exactly the same for the encoder and the decoder in order for the decoder to reproduce the original character sequence.

From the perspective of the component of a compression program which creates the predicted probability distributions, compression and decompression is equivalent. In order to generate a predicted probability distribution for a specific character, it takes as input all previous characters in the sequence. In the case of compression, it has access to these characters directly from the file it is trying to compress. In the case of decompression, it has access to these characters from the output of the arithmetic decoder. Since the predictor has to output the exact same probability distributions for both compression and decompression, any randomized components of the algorithm need to be initialized with the same seed.

\begin{figure}[t]
\vskip 0.1in
\begin{center}
\centerline{\includegraphics[width=3.5in]{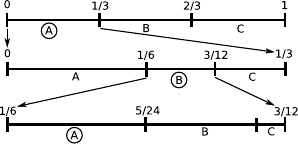}}
\caption{An example of arithmetic coding. The alphabet consists of three characters: A, B, and C. The string being encoded is ``ABA''.}
\label{coding}
\end{center}
\vskip -0.2in
\end{figure}

The process used to encode a sequence of characters by an arithmetic encoder is essentially equivalent to storing a single number (between 0 and 1). We will present how this process works through a small example. Suppose we have an alphabet of three characters: A, B, and C. Given the file ``ABA'' our goal is to compress this string using arithmetic encoding. For the first character prediction, assume that the predictor gives a uniform probability distribution. We can visualize this on a number line, as seen in the top of Figure~\ref{coding}. For the second character, assume that the predictor assigns a probability of 0.5 to A, 0.25 to B, and 0.25 to C. Since the first character in the sequence was A, the arithmetic encoder expands this region (0 to 1/3) and assigns the second predicted probability distribution according to this expanded region. This is visualized in the middle layer of Figure~\ref{coding}. For the final character in the sequence, assume that the predictor assigns a probability of 0.5 to A, 0.4 to B, and 0.1 to C. This is visualized in the bottom of Figure~\ref{coding}. Now all the arithmetic coder needs to do is store a single number between the values of 1/6 and 5/24. This number can be efficiently encoded using a binary search. The binary search ranges would be: ``0 to 1'', ``0 to 0.5'', ``0 to 0.25'', and finally ``0.125 to 0.25''. This represents the number 0.1875 (which falls in the desired range). If we use ``0'' to encode the decision to use the lower half and ``1'' to encode the decision to use the upper half, this sequence can be represented in binary as ``001''.

Now consider the task of decoding the file. As input, the arithmetic decoder has the number 0.1875, and a sequence of predicted probability distributions. For the first character, the predictor gives a uniform probability distribution. The number 0.1875 falls into the `A' sector, so the arithmetic decoder tells the predictor that the first character was `A'. Similarly, for the next two characters the arithmetic decoder knows that the characters must be `B' and `A'. At this point, the arithmetic decoder needs some way to know that it has reached the end of the sequence. There are typically two techniques that are used to communicate the length of the sequence to the decoder. The first is to encode a special ``end of sequence'' character, so that when the decoder reaches this character it knows it reached the end of the string. The second technique is to just store an additional integer along with the compressed file which represents the length of the sequence (this is usually more efficient in practice).

Although arithmetic coding can achieve optimal compression in theory, in practice there are two factors which prevent this. The first is the fact that files can only be stored to disk using a sequence of bytes, so this requires some overhead in comparison to storing the optimal number of bits. The second is the fact that precision limitations of floating point numbers prevent optimal encodings. In practice both of these factors result in relatively small overhead, so arithmetic coding still produces near-optimal encodings.

\subsection{PPM}
\label{PPM}

\begin{table*}[t]
\footnotesize{
\renewcommand{\arraystretch}{1.3}
\caption{{PPM} model after processing the string ``abracadabra'' (up to the second order model). This table is a recreation of a table from \citep{Cleary1995}.}
\vskip 0.15in
\label{context}
\begin{center}
\addtolength{\tabcolsep}{-1pt}
{ \begin{tabular}{|ccccc|ccccc|cccc|cccc|}
\hline
 \multicolumn{5}{|c|}{Order k = 2} &\multicolumn{5}{c|}{Order k = 1} &\multicolumn{4}{c|}{Order k = 0} &\multicolumn{4}{c|}{Order k = -1}\\
 
\multicolumn{3}{|c}{Predictions} & \multicolumn{1}{c}{c} &\multicolumn{1}{c}{p} & \multicolumn{3}{|c}{Predictions} & \multicolumn{1}{c}{c} &\multicolumn{1}{c}{p} & \multicolumn{2}{|c}{Predictions} & \multicolumn{1}{c}{c} &\multicolumn{1}{c}{p} & \multicolumn{2}{|c}{Predictions} & \multicolumn{1}{c}{c} &\multicolumn{1}{c|}{p} \\

\hline
ab&$\rightarrow$&r&2&$2 \over 3$&a&$\rightarrow$&b&2&$2 \over 7$&$\rightarrow$&a&5&$5 \over 16$&$\rightarrow$&A&1&$1 \over |A|$\\
&$\rightarrow$&Esc&1&$1 \over 3$&&$\rightarrow$&c&1&$1 \over 7$&$\rightarrow$&b&2&$2 \over 10$&&&&\\
&&&&&&$\rightarrow$&d&1&$1 \over 7$&$\rightarrow$&c&1&$1 \over 16$&&&&\\
ac&$\rightarrow$&a&1&$1 \over 2$&&$\rightarrow$&Esc&3&$3 \over 7$&$\rightarrow$&d&1&$1 \over 16$&&&&\\
&$\rightarrow$&Esc&1&$1 \over 2$&&&&&&$\rightarrow$&r&2&$2 \over 16$&&&&\\
&&&&&b&$\rightarrow$&r&2&$2 \over 3$&$\rightarrow$&Esc&5&$5 \over 16$&&&&\\
ad&$\rightarrow$&a&1&$1 \over 2$&&$\rightarrow$&Esc&1&$1 \over 3$&&&&&&&&\\
&$\rightarrow$&Esc&1&$1 \over 2$&&&&&&&&&&&&&\\
&&&&&c&$\rightarrow$&a&1&$1 \over 2$&&&&&&&&\\
br&$\rightarrow$&a&2&$2 \over 3$&&$\rightarrow$&Esc&1&$1 \over 2$&&&&&&&&\\
&$\rightarrow$&Esc&1&$1 \over 3$&&&&&&&&&&&&&\\
&&&&&d&$\rightarrow$&a&1&$1 \over 2$&&&&&&&&\\
ca&$\rightarrow$&d&1&$1 \over 2$&&$\rightarrow$&Esc&1&$1 \over 2$&&&&&&&&\\
&$\rightarrow$&Esc&1&$1 \over 2$&&&&&&&&&&&&&\\
&&&&&r&$\rightarrow$&a&2&$2 \over 3$&&&&&&&&\\
da&$\rightarrow$&b&1&$1 \over 2$&&$\rightarrow$&Esc&1&$1 \over 3$&&&&&&&&\\
&$\rightarrow$&Esc&1&$1 \over 2$&&&&&&&&&&&&&\\
&&&&&&&&&&&&&&&&&\\
ra&$\rightarrow$&c&1&$1 \over 2$&&&&&&&&&&&&&\\
&$\rightarrow$&Esc&1&$1 \over 2$&&&&&&&&&&&&&\\

\hline
\end{tabular}}
\end{center}}
\end{table*} 

{PPM} \citep{Cleary84} is a lossless compression algorithm which consistently
performs well on text compression benchmarks. It creates predicted probability distributions based on the history of characters in a sequence using a technique called context matching.

Consider the alphabet of lower case English characters and the input sequence ``abracadabra''. For
each character in this string, {PPM} needs to create a probability distribution representing how likely
the character is to occur. For the first character in the sequence, there is no prior information about
what character is likely to occur, so assigning a uniform distribution is the optimal strategy. For the
second character in the sequence, `a' can be assigned a slightly higher probability because it has
been observed once in the input history. Consider the task of predicting the character after the entire sequence. One way to go
about this prediction is to find the longest match in the input history which matches the most recent
input. In this case, the longest match is ``abra'' which occurs in the first and eighth
positions. Based on the longest match, a good prediction for the next
character in the sequence is simply the character immediately after the match in the input history. After the string ``abra'' was the character `c' in the fifth position. Therefore `c' is a good
prediction for the next character. Longer context matches can result in better predictions than shorter ones. This is because longer
matches are less likely to occur by chance or due to noise in the data. 

{PPM} essentially creates probability distributions according to the method described above. Instead
of generating the probability distribution entirely based on the longest context match, it blends the
predictions of multiple context lengths and assigns a higher weight to longer matches. There are
various techniques on how to go about blending different context lengths. The strategy used for combining different context lengths is partially responsible for the performance differences between various {PPM} implementations.

One example of a technique used to generate the predicted probabilities is shown in Table~\ref{context} \citep{Cleary1995}. The table shows the state of the model after the string ``abracadabra'' has been processed. `k' is the order of the context match, `c' is the occurence count for the context, and `p' is the computed probability. `Esc' refers to the event of an unexpected character and causes the algorithm to use a lower order model (weighted by the probability of the escape event). Note that the lowest order model (-1) has no escape event since it matches any possible character in the alphabet A.

{PPM} is a nonparametric model that adaptively changes based on the data it is compressing. It is not surprising that similar methods have been discovered in the field of Bayesian nonparametrics. The stochastic memoizer \citep{Wood09} is a nonparametric model based on an unbounded-depth hierarchical Pitman-Yor process. The stochastic memoizer shares several similarities with {PPM} implementations. The compression performance of the stochastic memoizer is currently comparable with some of the best {PPM} implementations.

\subsection{Compression Metrics}
\label{metrics}

One way of measuring compression performance is to use the file size of compressed data. However,
file size is dependent on a particular type of coding scheme (such as arithmetic coding or Huffman
coding). There are two common metrics used to measure the performance directly based on the predicted probability
distributions: cross entropy and perplexity. Cross entropy can be used to estimate
the average number of bits needed to code each byte of the original data. For a sequence of $N$
characters $x_i$, and a probability $p(x_i)$ assigned to each character by the prediction algorithm, the
cross entropy can be defined as:
$-\sum_{i=1}^N \frac{1}{N} \log_2 p(x_i)$.
This gives the expected number of bits needed to code each character of the string. Another common
metric used to compare text prediction algorithms is perplexity which can be defined as two to the
power of cross entropy.

\subsection{Lossless Image Compression}
\label{images}

Compressing images represents a significantly different problem than compressing text. Lossless 
compression algorithms tend to work best on a sequence of characters which contain relatively little noise. They
are well suited for natural language because the characters are highly redundant and contain less noise than individual pixels. The problem with noise is that it reduces the maximum context
lengths which an algorithm like {PPM} can identify. It
has been shown that a variant of {PPM} called prediction by partial approximate matching ({PPAM})
shows competitive compression rates when compared to other lossless image compression algorithms \citep{Zhang08}. {PPAM} uses approximate matches in order to find longer context lengths which can
improve the compression ratio for images.

\begin{figure}[t]
\vskip 0.1in
\begin{center}
\centerline{\includegraphics[width=2.8in]{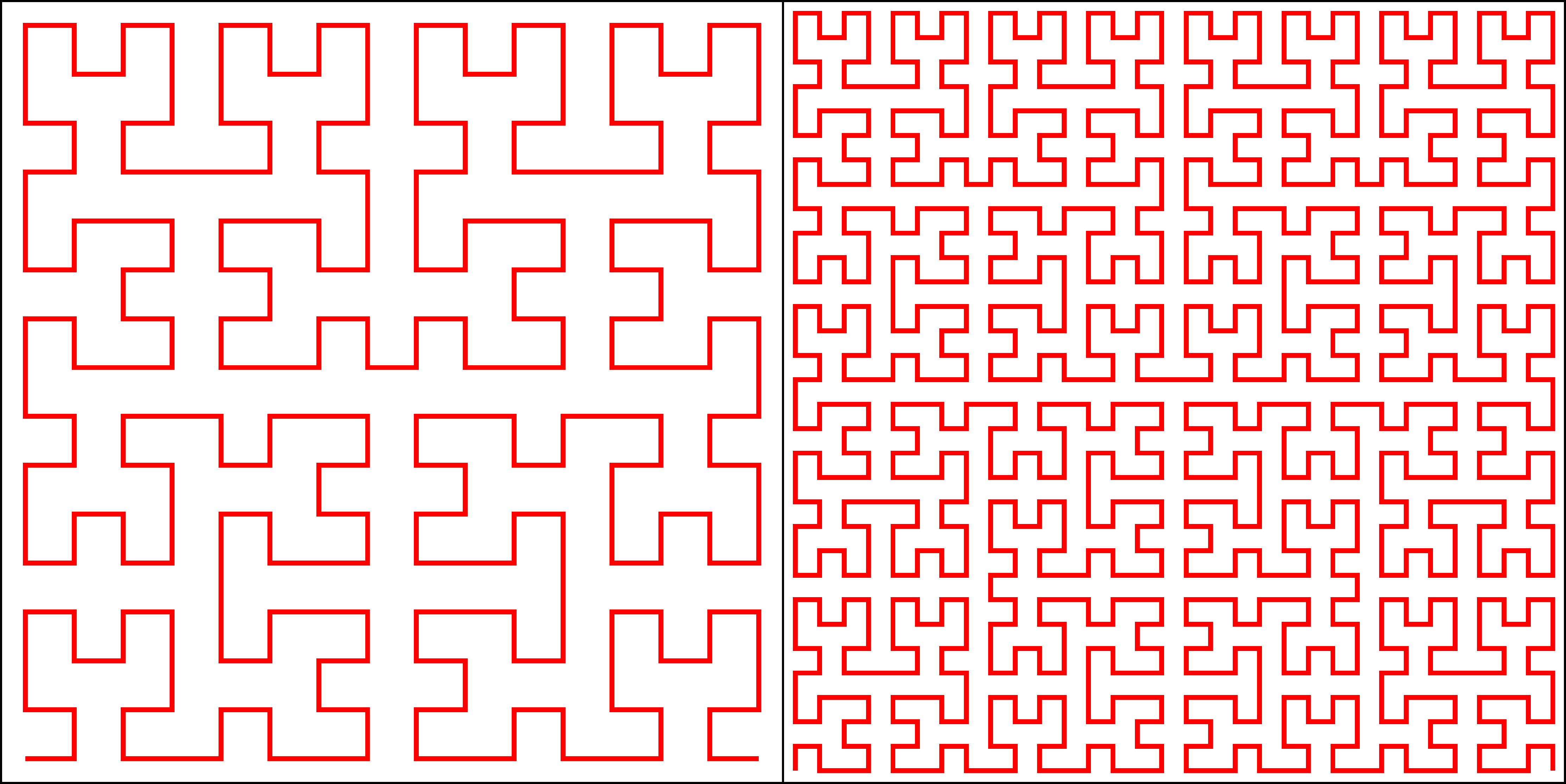}}
\caption{The fourth and fifth iteration of the Hilbert curve construction. Image courtesy of Zbigniew Fiedorowicz.}
\label{timefigure}
\end{center}
\vskip -0.2in
\end{figure}

Another fundamental difference between text and images is that text is a single dimensional sequence of characters while images are two dimensional. Applying {PPM} to image
compression requires mapping the pixels to a single sequential dimension. A trivial way of performing this mapping is using a raster scan ($i.e.$ scanning rows from top to bottom and pixels in each row from left to right). Another mapping known as the Hilbert curve \citep{Hilbert1891} maximizes spatial locality (shown in Figure~\ref{timefigure}).

\subsection{Distance Metrics}
\label{metric}

Keogh et~al \citep{Keogh04} demonstrate how compression algorithms can be used as a distance metric between time series data. This distance metric can be used to solve several interesting problems such as clustering, anomaly detection, and classification. For example, the distance metric can be used to cluster a variety of types of files such as music, text documents, images, and genome sequences. Li et~al \citep{Li01} propose the following metric to measure the distance between the strings $x$ and $y$:
\bea
d_{k}(x,y) = {{K(x|y) + K(y|x)} \over {K(xy)}}\nonumber
\eea
$K(x)$ is defined to be the Kolmogorov complexity of a string $x$. That is the length of the shortest program capable of producing $x$ on a universal computer. $K(x|y)$ is the length of the shortest program that computes $x$ when $y$ is given as auxiliary input to the program. $K(xy)$ is the length of the shortest program that outputs $y$ concatenated to $x$. Kolmogorov complexity represents the best possible compression that can be achieved. Since the Kolmogorov complexity can not be directly computed, a compression algorithm can be used to approximate $d_{k}(x,y)$:
\bea
d_{c}(x,y) = {{C(x|y) + C(y|x)} \over {C(xy)}}\nonumber
\eea
$C(x)$ is the compressed size of $x$ and $C(x|y)$ is the compressed size of $x$ after the compressor has been trained on $y$. The better a compression algorithm, the closer it approaches the Kolmogorov complexity and the closer $d_{c}(x,y)$ approximates $d_{k}(x,y)$. Keogh et~al \citep{Keogh04} use the following dissimilarity metric: 
\bea
d_{CDM}(x,y) = {{C(xy)} \over {C(x) + C(y)}}\nonumber
\eea
The main justification made for using $d_{CDM}(x,y)$ over $d_{c}(x,y)$ is that it does not require the calculation of $C(x|y)$ and $C(y|x)$. These can not be calculated using most off-the-shelf compression algorithms without modifying their source code. Fortunately, PAQ8 is open source and these modifications can be easily implemented (so $d_{c}$ can be used). For the purposes of classification, we investigated defining our own distance metrics. Using cross entropy, a more computationally efficient distance metric can be defined which requires only one pass through the data:
\bea
d_{e1}(x,y) = E(x|y) \nonumber
\eea
$E(x|y)$ is the cross entropy of $x$ after the compressor has been trained on $y$. We also investigated a symmetric version of this distance metric, in which $d_{e2}(x,y)$ is always equal to $d_{e2}(y,x)$:
\bea
d_{e2}(x,y) = {{E(x|y) + E(y|x)} \over {2}}\nonumber
\eea
Finally, Cilibrasi et~al \citep{Cilibrasi2005} propose using the following distance metric:
\bea
d_{NDM}(x,y) = {{C(xy) - min(C(x),C(y))} \over {max(C(x),C(y))}}\nonumber
\eea
In section~\ref{ch:Classification} of this report, we use a compression-based distance metric to perform classification. Keogh et~al \citep{Keogh04} use the ZIP compression algorithm as a distance metric to perform experiments in clustering, anomaly detection, and classification. Since PAQ8 achieves better compression than ZIP, it should theoretically result in a better distance metric. Although we do not perform the experiments in this report, it would make interesting future work to compare ZIP and PAQ8 in the experiments by Keogh et~al.


\section{PAQ8}
\label{ch:PAQ8}

\subsection{Architecture}

PAQ8 uses a weighted combination of predictions from a large number of models. Most of the models are based on context matching. Unlike {PPM}, some of the models allow noncontiguous context matches. Noncontiguous context matches improve noise robustness in comparison to {PPM}. This also enables PAQ8 to capture longer-term dependencies. Some of the models are specialized for particular types of data such as images or spreadsheets. Most {PPM} implementations make predictions on the byte-level (given a sequence of bytes, they predict the next byte). However, all of the models used by PAQ8 make predictions on the bit-level. 

Some architectural details of PAQ8 depend on the version used. Even for a particular version of PAQ8, the algorithm changes based on the type of data detected. For example, fewer prediction models are used when image data is detected. We will provide a high-level overview of the architecture used by {\tt paq8l} in the general case of when the file type is not recognized. {\tt paq8l} is a stable version of PAQ8 released by Matt Mahoney in March 2007. The PAQ8 versions submitted to the Hutter prize include additional language modeling components not present in {\tt paq8l} such as dictionary preprocessing and word-level modeling.

\begin{figure}[ht]
\vskip 0.1in
\begin{center}
\centerline{\includegraphics[width=3.0in]{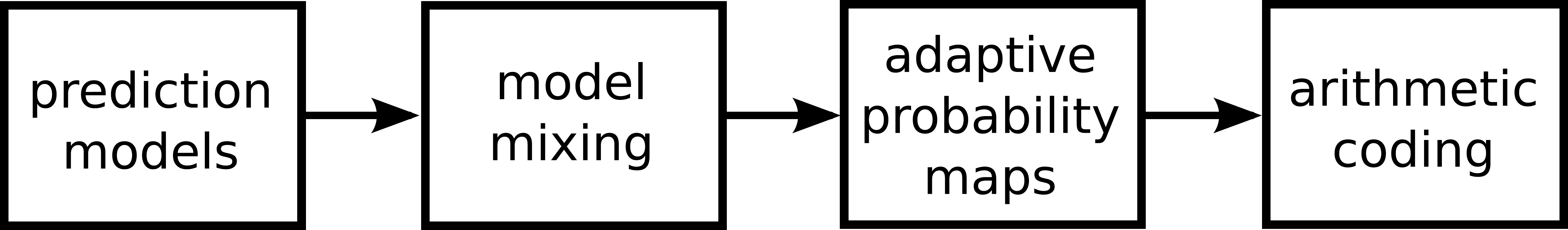}}
\caption{PAQ8 architecture.}
\label{paq8-figure}
\end{center}
\vskip -0.2in
\end{figure}

An overview of the {\tt paq8l} architecture is shown in Figure~\ref{paq8-figure}. 552 prediction models are used. The model mixer combines the output of the 552 predictors into a single prediction. This prediction is then passed through an adaptive probability map (APM) before it is used by the arithmetic coder. In practice, APMs typically reduce prediction error by about 1\%. APMs are also known as secondary symbol estimation \citep{Mahoney11}. APMs were originally developed by Serge Osnach for PAQ2. An APM is a two dimensional table which takes the model mixer prediction and a low order context as inputs and outputs a new prediction on a nonlinear scale (with finer resolution near 0 and 1). The table entries are adjusted according to prediction error after each bit is coded.

\subsection{Model Mixer}

\begin{figure*}[ht]
\vskip 0.1in
\begin{center}
\centerline{\includegraphics[width=5.2in]{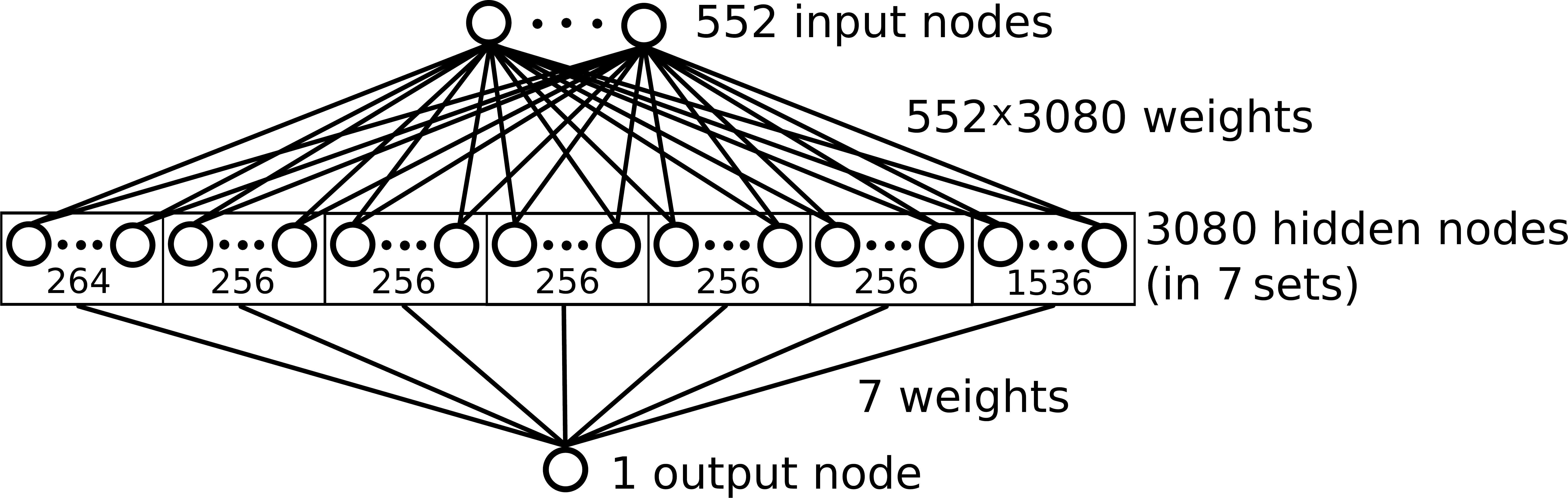}}
\caption{PAQ8 model mixer architecture.}
\label{paq8-mixer}
\end{center}
\vskip -0.2in
\end{figure*}

The {\tt paq8l} model mixer architecture is shown in Figure~\ref{paq8-mixer}. The architecture closely resembles a neural network with one hidden layer. However, there are some subtle differences that distinguish it from a standard neural network. The first major difference is that weights for the first and second layers are learned online and independently for each node. Unlike back propagation for a multi-layer network, each node is trained separately to minimize the predictive cross-entropy error, as outlined in section~\ref{update}. In this sense, PAQ8 is a type of ensemble method \citep{Opitz99}. Unlike typical ensembles, the parameters do not converge to fixed values unless the data is stationary. PAQ8 was designed for both stationary and non-stationary data\footnote[1]{We refer to ``non-stationary data'' as data in which the statistics change over time. For example, we would consider a novel to be non-stationary while a text document of some repeating string ($e.g.$ ``abababab...'') to be stationary.}.

The second major difference between the model mixer and a standard neural network is the fact that the hidden nodes are partitioned into seven sets. For every bit of the data file, one node is selected from each set. The set sizes are shown in the rectangles of Figure~\ref{paq8-mixer}. We refer to the leftmost rectangle as set 1 and the rightmost rectangle as set 7. Only the edges connected to these seven selected nodes are updated for each bit of the data. That means of the 552$\times$3,080 = 1,700,160 weights in the first layer, only 552$\times$7 = 3,864 of the weights are updated for each bit. This makes training the neural network several orders of magnitude faster.

Each set uses a different selection mechanism to choose a node. Sets number 1, 2, 4, and 5 choose the node index based on a single byte in the input history. For example, if the byte for set 1 has a value of 4, the fifth node of set 1 would be selected. Set 1 uses the second most recent byte from the input history, set 2 uses the most recent byte, set 4 uses the third most recent byte, and set 5 uses the fourth most recent byte. Set 6 chooses the node based on the length of the longest context matched with the most recent input. Sets 3 and 7 use a combination of several bytes of the input history in order to choose a node index.
The selection mechanism used by {\tt paq8l} is shown in Algorithm~\ref{tab:code}. $history(i)$ returns the $i$'th most recent byte, $lowOrderMatches$ the number of low-order contexts which have been observed at least once before (between 0 and 7), $lastFourBytes$ is the four most recent bytes, $longestMatch$ is the length of the longest context match (between 0 and 65534), $bitMask(x,y)$ does a bitwise AND operation between $x$ and $y$, and $bitPosition$ is the bit index of the current byte (between 0 and 7).

\begin{algorithm}[t]
\caption{{\tt paq8l} node selection mechanism.}
\label{tab:code}
\begin{algorithmic}
\STATE \footnotesize set1Index$\gets$ 8 + history(1)
\STATE set2Index$\gets$ history(0)
\STATE set3Index$\gets$ lowOrderMatches + 8 $\times$ ((lastFourBytes/32) mod (8))
\IF {history(1) = history(2)}
	\STATE set3Index$\gets$ set3Index + 64
\ENDIF
\STATE set4Index$\gets$ history(2)
\STATE set5Index$\gets$ history(3)
\STATE set6Index$\gets$ round($\log_{2}$(longestMatch) $\times$ 16)

\IF {$bitPosition = 0$}
	\STATE set7Index$\gets$ history(3)/128 + bitMask(history(1),240) + 4$\times$(history(2)/64) + 2$\times$(lastFourBytes / $2^{31}$)
\ELSE
	\STATE set7Index$\gets$ history(0) $\times 2^{8-\mathrm{bitPosition}}$
	\IF {bitPosition = 1}
		\STATE set7Index$\gets$ set7Index + history(3)/2
	\ENDIF
	\STATE set7Index$\gets$ $\min$(bitPosition,5) $\times$ 256 + history(1)/32 + 8 $\times$ (history(2)/32) + bitMask(set7Index,192)
\ENDIF
\end{algorithmic}
\end{algorithm}

\subsubsection{Mixtures of Experts}

\begin{figure}[ht]
\vskip 0.1in
\begin{center}
\centerline{\includegraphics[width=3.2in]{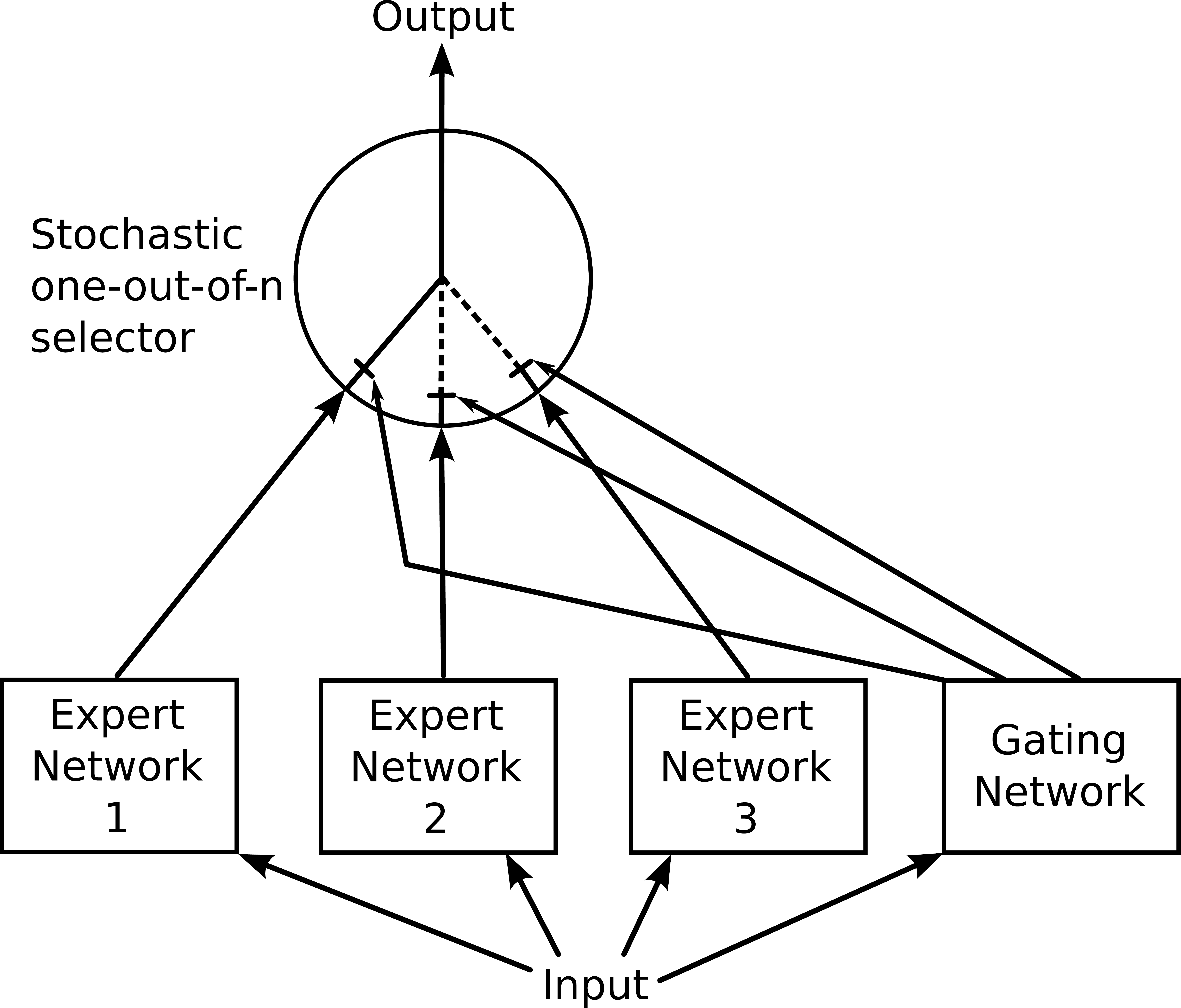}}
\caption{Mixtures of experts architecture. This figure is a recreation of a figure in \citep{Jacobs1991}. All of the experts are feedforward networks and have the same input. The gating network acts as a switch to select a single expert. The output of the selected expert becomes the output of the system. Only the weights of the selected expert are trained.}
\label{fig:experts}
\end{center}
\vskip -0.2in
\end{figure}

In the previous section we compared the PAQ8 model mixer to a multilayer neural network. The PAQ8 model mixer can also be compared to a technique known as ``mixtures of experts'' \citep{Jacobs1991}. Although PAQ8 does not use the standard mixtures of experts architecture, they do share some similarities. Jacobs et~al \citep{Jacobs1991} state: ``If backpropagation is used to train a single, multilayer network to perform different subtasks
on different occasions, there will generally be strong interference effects which lead to slow
learning and poor generalization. If we know in advance that a set of training cases may
be naturally divided into subsets that correspond to distinct subtasks, interference can be
reduced by using a system composed of several different `expert' networks plus a gating
network that decides which of the experts should be used for each training case.'' Their architecture is shown in Figure~\ref{fig:experts}.

PAQ8 and the mixtures of experts architecture both use a gating mechanism to choose expert models. The same problem-specific properties which lead to the development of mixtures of experts also applies to compression - data can be naturally divided into subsets and separate `experts' can be trained on each subset. Using a gating mechanism has the additional computational benefit that only one expert needs to be trained at a time, instead of training all experts simultaneously. Increasing the number of expert networks does not increase the time complexity of the algorithm.

One difference between the mixtures of experts model and the PAQ8 model mixture is the gating mechanism. Jacobs et~al use a feedforward network to learn the gating mechanism, while PAQ8 uses a deterministic algorithm (shown in Algorithm~\ref{tab:code}) which does not perform adaptive learning. The gating algorithm used by PAQ8 contains problem-specific knowledge which is specified a priori. One interesting area for future work would be to investigate the effect of adaptive learning in the PAQ8 gating mechanism. Adaptive learning could potentially lead to a better distribution of the data to each expert. Ideally, the data should be uniformly partitioned across all the experts. However, using a deterministic gating mechanism runs the risk of a particular expert being selected too often.

The gating mechanism in PAQ is governed by the values of the input data. The idea of gating units in a network according to the value of the input has also been used in other recurrent neural network architectures. For example, in long-short-term-memory (LSTM), it is used to maintain hidden units switched-on and hence avoid the problem of vanishing gradients in back-propagation, see e.g. \cite{Graves-2009-lstm}. However, the deterministic gating mechanism of PAQ is not intended at improving prediction performance or avoiding vanishing gradients. Rather, its objective is to reduce computation vastly. We recommend that researchers working with RNNs, which take days to train, investigate ways of incorporating these ideas to speed up the training of RNNs. Adaptive training, instead of training to a fixed value, is another important aspect to keep in mind.

There is a direct mapping between the mixtures of experts architecture and the PAQ8 model mixer architecture. Each ``set'' in the hidden layer of Figure~\ref{paq8-mixer} corresponds to a separate mixture of experts model. The seven mixtures of experts are then combined using an additional feedforward layer. The number of experts in each mixtures of experts model corresponds to the number of nodes in each set (\emph{e.g.} there are 264 experts in set 1 and 1536 experts in set 7). As with the mixtures of experts architecture, only the weights of the expert chosen by the gating mechanism are trained for each bit of the data. Another difference between the standard mixtures of experts model and PAQ8 is the fact that mixtures of experts models typically are optimized to converge towards a stationary objective function while PAQ8 is designed to adaptively train on both stationary and non-stationary data.

\subsubsection{Online Parameter Updating}
\label{update}

Each node of the {\tt paq8l} model mixer (both hidden and output) is a Bernoulli logistic model:
\be
p(y_t|\vx_t,\vw) = \Ber(y_t | \sigmoid(\vw^T \vx_t)) \nonumber
\ee
where $\vw \in\Bbb{R}^{n_p}$ is the vector of weights, $\vx_t\in [0,1]^{n_p}$ is the vector of predictors at time $t$, $y_t \in \{0,1\}$ is the next bit in the data being compressed, and $\sigmoid(\eta) = 1/(1+e^{-\eta})$ is the sigmoid or logistic function. $n_p$ is the number of predictors and is equal to 552 for the first layer of the neural network and 7 for the second layer of the network.
Let $\pi_t = \sigmoid(\vw^T \vx_t)$. The negative log-likelihood of the $t$-th bit
is given by
\bea
NLL(\vw) &=&
  -\log[ \pi_t^{\ind{y_t=1}} \times (1-\pi_t)^{\ind{y_t=0}}] = - \left[ y_t \log \pi_t + (1-y_t) \log (1-\pi_t) \right]
\nonumber
\eea
where $\ind{\cdot}$ denotes the indicator function. The last expression is the cross-entropy error (also known as coding error) function term at time $t$. The logistic regression weights are updated online with first order updates:
\[
\vw_{t} = \vw_{t-1} - \eta \nabla NLL(\vw_{t-1})
= \vw_{t-1} - \eta (\pi_t - y_t) \vx_t
\]
The step size $\eta$ is held constant to ensure ongoing adaptation.

\subsubsection{Extended Kalman Filter}

To improve the compression rate of {\tt paq8l}, we applied an extended Kalman filter ({EKF}) to adapt the weights. We assume a dynamic state-space model consisting of a Gaussian transition prior, $\vw_{t+1}=\vw_{t}+{\cal N}(0,\vQ)$, and a logistic observation model $y_t = \pi_t + {\cal N}(0,r)$.
The {EKF}, although based on local linearization of the observation model, is a second order adaptive method worthy of investigation. One of the earliest implementations of {EKF} to train multilayer perceptrons is due to Singhal and Wu \citep{Singhal1989}. Since {EKF} has a $O({n_p}^2)$ time complexity, it would be unfeasibly slow to apply {EKF} to the first layer of the neural network. However, {EKF} can be used to replace the method used by {\tt paq8l} in the second layer of the neural network without significant computational cost since there are only seven weights.

Here we present the {EKF} algorithm for optimizing the second layer of the PAQ8 neural network. The following values were used to initialize {EKF}:
$\vQ = 0.15\times \vI_{\scriptscriptstyle 7\times 7}$,
$\vP_0 = 60\times \vI_{\scriptscriptstyle 7\times 7}$,
$\vw_{0} = 150 \times \textbf{1}_{\scriptscriptstyle 7\times 1}$,
and $r = 5$.
The following are the {EKF} update equations for each bit of data:
\bea
\vw_{t+1|t} &=& \vw_t \nonumber\\
\vP_{t+1|t} &=& \vP_t + \vQ \nonumber\\
\vK_{t+1} &=& {\vP_{t+1|t}\vG'_{t+1}} \over {r + \vG_{t+1}\vP_{t+1|t}\vG'_{t+1}} \nonumber\\
\vw_{t+1} &=& \vw_{t+1|t} + \vK_{t+1} (y_{t}-\pi_{t})\nonumber\\
\vP_{t+1} &=& \vP_{t+1|t} - \vK_{t+1}\vG_{t+1}\vP_{t+1|t},\nonumber
\eea
where $\textbf{G}_{\scriptscriptstyle 1\times 7}$ is the Jacobian matrix:
$
\vG = \left[\partial y / \partial w_1 \;\; \cdots \;\;  {\partial y / \partial w_7}\right]$ with $
{\partial y / \partial w_i} = y(1-y)x_i.
$
We compared the performance of {EKF} with other variants of {\tt paq8l}. The results are shown in Table~\ref{versions}. The first three columns are {\tt paq8l} with different settings of the \emph{level} parameter. \emph{level} is the only {\tt paq8l} parameter that can be changed via command-line (without modifying the source code). It makes a tradeoff between speed, memory usage, and compression performance. It can be set to an integer value between zero and eight. Lower values of \emph{level} are faster and use less memory but achieve worse compression performance. \emph{level}=8 is the slowest setting and uses the most memory (up to 1643 MiB) but achieves the best compression performance. \emph{level}=5 has a 233 MiB memory limit. {\tt paq8-8-tuned} is a customized version of {\tt paq8l} (with \emph{level}=8) in which we changed the value of the weight initialization for the second layer of the neural network. We found changing the initialization value from 32,767 to 128 improved compression performance. Finally, {\tt paq8-8-ekf} refers to our modified version of {\tt paq8l} with {EKF} used to update the weights in the second layer of the neural network. We find that using {EKF} slightly outperforms the first order updates. The improvement is about the same order of magnitude as the improvement between \emph{level}=5 and \emph{level}=8. However, changing \emph{level} has a significant cost in memory usage, while using {EKF} has no significant computational cost. The initialization values for {\tt paq8-8-tuned} and {\tt paq8-8-ekf} were determined using manual parameter tuning on the first Calgary corpus file (`bib'). The performance difference between {\tt paq8l-8} and {\tt paq8-8-tuned} is similar to the difference between {\tt paq8-8-tuned} and {\tt paq8-8-ekf}.

\begin{table*}[t]
\caption{PAQ8 compression rates on the Calgary corpus}
\label{versions}
\vskip 0.15in
\begin{center}
\begin{small}
\begin{sc}
{ \begin{tabular}{lccccc}
\hline
\abovespace\belowspace
File & {\tt paq8l-1} & {\tt paq8l-5} & {\tt paq8l-8} & {\tt paq8-8-tuned} & {\tt paq8-8-ekf} \\
\hline
\abovespace
bib & 1.64592 & 1.49697 & 1.49645 & 1.49486 & 1.49207 \\
book1 & 2.14158 & 2.00573 & 2.00078 & 2.00053 & 1.99603 \\
book2 & 1.73257 & 1.59531 & 1.5923 & 1.59198 & 1.58861 \\
geo & 3.70451 & 3.43456 & 3.42725 & 3.42596 & 3.43444 \\
news & 2.07839 & 1.90573 & 1.90284 & 1.90237 & 1.89887 \\
obj1 & 3.25932 & 2.77358 & 2.77407 & 2.76531 & 2.76852 \\
obj2 & 1.85614 & 1.45499 & 1.43815 & 1.43741 & 1.43584 \\
paper1 & 2.09455 & 1.96543 & 1.96542 & 1.96199 & 1.95753 \\
paper2 & 2.09389 & 1.99046 & 1.99053 & 1.9882 & 1.98358 \\
pic & 0.6604 & 0.35088 & 0.35083 & 0.35073 & 0.3486 \\
progc & 2.07449 & 1.91574 & 1.91469 & 1.91037 & 1.9071 \\
progl & 1.31293 & 1.18313 & 1.18338 & 1.1813 & 1.18015 \\
progp & 1.31669 & 1.1508 & 1.15065 & 1.14757 & 1.14614 \\
trans & 1.1021 & 0.99169 & 0.99045 & 0.98857 & 0.98845 \\
\belowspace
{\bf Average} & {\bf1.93382} & {\bf1.72964} & {\bf1.72698} & {\bf1.7248} & {\bf1.72328} \\
\hline
\end{tabular}}
\end{sc}
\end{small}
\end{center}
\vskip -0.1in
\end{table*}


\section{Applications}
\label{ch:Applications}

\subsection{Adaptive Text Prediction and Game Playing}
\label{sec:game}
The fact that PAQ8 achieves state of the art compression results on text documents indicates that it can be used as a powerful model for natural language. PAQ8 can be used to find the string $x$ that maximizes $p(x|y)$ for some training string $y$. It can also be used to estimate the probability $p(z|y)$ of a particular string $z$ given some training string $y$. Both of these tasks are useful for several natural language applications. For example, many speech recognition systems are composed of an acoustic modeling component and a language modeling component. PAQ8 could be used to directly replace the language modeling component of any existing speech recognition system to achieve more accurate word predictions.

Text prediction can be used to minimize the number of keystrokes required to type a particular string \citep{Vitoria2006}. These predictions can be used to improve the communication rate for people with disabilities and for people using slow input devices (such as mobile phones). We modified the source code of {\tt paq8l} to create a program which predicts the next $n$ characters while the user is typing a string. A new prediction is created after each input character is typed. It uses {\tt fork()} after each input character to create a process which generates the most likely next $n$ characters. {\tt fork()} is a system call on Unix-like operating systems which creates an exact copy of an existing process. The program can also be given a set of files to train on.

Some preliminary observational studies on our text prediction system are shown in Figure~\ref{fig:prediction}. Note that PAQ8 continuously does online learning, even while making a prediction (as seen by the completion of ``Byron Knoll'' in the top example). The character predictions do capture some syntactic structures (as seen by completion of LaTeX syntax) and even some semantic information as implied by the training text.

\begin{figure}[t]
\vskip 0.1in

{\scriptsize

\noindent
{\bf M}|ay the contemplation of so many wonders extinguish the spirit ofvengeance in him!

\noindent
{\bf My}| companions and I had decided to escape as soon as the vessel cameclose enough for us to be heard

\noindent
{\bf My n}|erves calmed a little, but with my brain so aroused,I did a swift review of my whole existence

\noindent
{\bf My name i}|n my ears and some enormous baleen whales

\noindent
{\bf My name is B}|ay of Bengal, the seas of the East Indies, the seasof China

\noindent
{\bf My name is Byr}|on and as if it was an insane idea.  But where the lounge.I stared at the ship bearing

\noindent
{\bf My name is Byron K}|eeling Island disappeared below the horizon,

\noindent
{\bf My name is Byron Kn}|ow how the skiff escaped from the Maelstrom'sfearsome eddies,

\noindent
{\bf My name is Byron Knoll.}|  It was an insane idea.  Fortunately I controlled myselfand stretched

\noindent
{\bf My name is Byron Knoll. My name is B}|yron Knoll. My name is Byron Knoll. My name is Byron Knoll.

\vspace{0.2cm}

\noindent
{\bf F}|or example, consider the form of the exponential family 

\noindent
{\bf Fi}|gure$\sim$\textbackslash ref\textbraceleft fig:betaPriorPost\textbraceright (c) shows what happens as the number of heads in the past data.

\noindent
{\bf Figure o}|f the data, as follows: \textbackslash bea\textbackslash gauss(\textbackslash mu$\mid$\textbackslash gamma, \textbackslash lambda(2 \textbackslash alpha-1))

\noindent
{\bf Figure ou}|r conclusions are a convex combination of the prior mean and the constraints 

\noindent
{\bf Figure out Bayesian theory we must. Jo}|rdan conjugate prior

\noindent
{\bf Figure out Bayesian theory we must. Jos}|h Tenenbaum point of the posterior mean is and mode of the

\noindent
{\bf Figure out Bayesian theory we must. Josh agrees. Long live P(\textbackslash vtheta}|$\mid$\textbackslash data)

}

\vspace{-2mm}
\caption{Two examples of PAQ8 interactive text prediction sessions. The user typed the text in boldface and PAQ8 generated the prediction after the ``|'' symbol. We shortened some of the predictions for presentation purposes. In the top example, PAQ8 was trained on ``Twenty Thousand Leagues Under the Seas'' (Jules Verne, 1869). In the bottom example, PAQ8 was trained on the LaTeX source of a machine learning book by Kevin P. Murphy.}
\label{fig:prediction}
\end{figure}

Sutskever et~al \citep{Sutskever11} use Recurrent Neural Networks ({{RNN}s) to perform text prediction. They also compare {{RNN}s to the sequence memoizer and PAQ8 in terms of compression rate. They conclude that {{RNN}s achieve better compression than the sequence memoizer but worse than PAQ8. They perform several text prediction tasks using {{RNN}s with different training sets (similar to the examples in Figure~\ref{fig:prediction}). One difference between their method and ours is the fact that PAQ8 continuously does online learning on the test data. This feature could be beneficial for text prediction applications because it allows the system to adapt to new users and data that does not appear in the training set.

We found that the PAQ8 text prediction program could be modified into a rock-paper-scissors AI that usually beats human players. Given a sequence of the opponent's rock-paper-scissors moves (such as ``rpprssrps'') it predicts the most likely next move for the opponent. In the next round the AI would then play the move that beats that prediction. The reason that this strategy usually beats human players is that humans typically use predictable patterns after a large number of rock-paper-scissors rounds. The PAQ8 text prediction program and rock-paper-scissors AI are available to be downloaded (see Appendix~\ref{ch:a}).

\subsection{Classification}
\label{ch:Classification}

In many classification settings of practical interest, the data appears in sequences (\emph{e.g.} text). Text categorization has particular relevance for classification tasks in the web domain (such as spam filtering). Even when the data does not appear to be obviously sequential in nature (\emph{e.g.} images), one can sometimes find ingenious ways of mapping the data to sequences.

Compression-based classification was discovered independently by several researchers \citep{Marton05}. One of the main benefits of compression-based methods is that they are very easy to apply as they usually require no data preprocessing or parameter tuning. There are several standard procedures for performing compression-based classification. These procedures all take advantage of the fact that when compressing the concatenation of two pieces of data, compression programs tend to achieve better compression rates when the data share common patterns. If a data point in the test set compresses well with a particular class in the training set, it likely belongs to that class. Any of the distance metrics defined in Section~\ref{metric} can be directly used to do classification (for example, using the k-nearest neighbor algorithm). We developed a classification algorithm using PAQ8 and show that it can be used to achieve competitive classification rates in two disparate domains: text categorization and shape recognition.

\subsubsection{Techniques}

Marton et~al \citep{Marton05} describe three common compression-based classification procedures: standard minimum description length ({SMDL}), approximate minimum description length ({AMDL}), and best-compression neighbor ({BCN}). Suppose each data point in the training and test sets are stored in separate files. Each file in the training set belongs to one of the classes \(C_1,...,C_N\). Let the file \(A_i\) be the concatenation of all training files in class \(C_i\). {SMDL} runs a compression algorithm on each \(A_i\) to obtain a model (or dictionary) \(M_i\). Each test file \(T\) is compressed using each \(M_i\). \(T\) is assigned to the class \(C_i\) whose model \(M_i\) results in the best compression of \(T\). While compressing $T$, the compression algorithm does not update the model $M_i$.

For a file \(F\), let \(f(F)\) be the file size of the compressed version of \(F\). Also, let \(A_iT\) be the file \(A_i\) concatenated with \(T\). For {AMDL}, \(T\) is assigned to the class \(C_i\) which minimizes the difference \(f(A_iT)\) - \(f(A_i)\). Let \(B\) be a file in the training set. {BCN} checks every pair \(BT\) and assigns \(T\) to the class which minimizes the difference \(f(BT)\) - \(f(B)\).

\begin{table}[t]
\caption{The number of times each bit of data gets compressed using different compression-based classification methods. $N_X$ is the number of training files, $N_Y$ is the number of test files, and $N_Z$ is the number of classes.}
\label{bits}
\vskip 0.15in
\begin{center}
\begin{small}
\begin{sc}
{ \begin{tabular}{lcccr}
\hline
\abovespace\belowspace
Method & Training data & Test data \\
\hline
\abovespace
SMDL	& $1$ & $N_Z$ \\
AMDL	& $N_{Y}+1$ & $N_Z$ \\
\belowspace
BCN	& $N_{Y}+1$ & $N_X$ \\
\hline
\end{tabular}}
\end{sc}
\end{small}
\end{center}
\vskip -0.1in
\end{table}

A speed comparison between these methods can be made by considering how many times each bit of the data gets compressed, as shown in Table~\ref{bits}. It should be noted that the primary difference between {SMDL} and {AMDL} is the fact that {SMDL} only processes the training data once, while {AMDL} reprocesses the training data for every file in the test set. For many datasets, the number of training and test files (\(N_X\) and \(N_Y\)) are much larger than the number of classes (\(N_Z\)). That means that {SMDL} can be orders of magnitude faster than {AMDL} (and {AMDL} faster than {BCN}). Although PAQ8 achieves state of the art compression rates, it is also extremely slow compared to the majority of compression algorithms. Using PAQ8 for classification on large datasets would be unfeasibly slow using {AMDL} or {BCN}.

\subsubsection{PAQclass}

{AMDL} and {BCN} both work with off-the-shelf compression programs. However, implementing {SMDL} usually requires access to a compression program's source code. Since PAQ8 is open source, we modified the source code of {\tt paq8l} (a version of PAQ8) to implement {SMDL}. We call this classifier PAQclass. To the best of our knowledge, PAQ has never been modified to implement {SMDL} before. We changed the source code to call {\tt fork()} when it finishes processing data in the training set (for a particular class). One forked process is created for every file in the test set. This essentially copies the state of the compressor after training and allows each test file to be compressed independently. Note that this procedure is slightly different from {SMDL} because the model \(M_i\) continues to be adaptively modified while it is processing test file \(T\). However, it still has the same time complexity as {SMDL}. {\tt paq8l} has one parameter to set the compression level. We used the default parameter setting of 5 during classification experiments.

Compression performance for {AMDL} and {BCN} is measured using file size. The use of file size is fundamentally limited in two ways. The first is that it is only accurate to within a byte (due to the way files are stored on disk). The second is that it is reliant on the non-optimal arithmetic coding process to encode files to disk. Cross entropy is a better measurement of compression performance because it is subject to neither of these limitations. Since we had access to the {\tt paq8l} source code, we used cross entropy as a measure of compression performance instead of file size.

\subsubsection{Text Categorization}

Text categorization is the problem of assigning documents to categories based on their content. We evaluated PAQclass on the 20 Newsgroup (20news) dataset. This dataset contains 18,828 newsgroup documents partitioned (nearly) evenly across 20 categories. The number of documents in each category is shown in Table~\ref{news-classes}. We used J. Rennie's version of the corpus, which is available at: \emph{http://people.csail.mit.edu/people/jrennie/20Newsgroups/20news-18828.tar.gz}. In this version of the corpus duplicate postings to more than one newsgroup were removed. Most message headers were also removed, while the ``Subject'' and ``From'' fields were retained.

\begin{table}[t]
\caption{Number of documents in each category of the 20news dataset.}
\label{news-classes}
\vskip 0.15in
\begin{center}
\begin{small}
\begin{sc}
\begin{tabular}{lcccr}
\hline
\abovespace\belowspace
Class & Count \\
\hline
\abovespace
alt.atheism			& 799 \\
comp.graphics			& 973 \\
comp.os.ms-windows.misc			& 985 \\
comp.sys.ibm.pc.hardware			& 982 \\
comp.sys.mac.hardware			& 961 \\
comp.windows.x			& 980 \\
misc.forsale			& 972 \\
rec.autos			& 990 \\
rec.motorcycles			& 994 \\
rec.sport.baseball			& 994 \\
rec.sport.hockey			& 999 \\
sci.crypt			& 991 \\
sci.electronics			& 981 \\
sci.med			& 990 \\
sci.space			& 987 \\
soc.religion.christian			& 997 \\
talk.politics.guns			& 910 \\
talk.politics.mideast			& 940 \\
talk.politics.misc			& 775 \\
talk.religion.misc			& 628 \\
\belowspace
{\bf Total} & {\bf 18,828} \\
\hline
\end{tabular}
\end{sc}
\end{small}
\end{center}
\vskip -0.1in
\end{table}

\begin{table}[t]
\caption{Classification results on the 20news dataset. Each row shows one run of a randomized 80-20 train-test split.}
\label{news}
\vskip 0.15in
\begin{center}
\begin{small}
\begin{sc}
{ \begin{tabular}{lcccr}
\hline
\abovespace
Correct classifications & Percent correct \\
\belowspace
(out of 3766) & \\
\hline
\abovespace
3470 & 92.1402 \\
3482 & 92.4588 \\
3466 & 92.034 \\
3492 & 92.7244 \\
3480 & 92.4057 \\
\belowspace
{\bf Average} & {\bf 92.3526} \\
\hline
\end{tabular}}
\end{sc}
\end{small}
\end{center}
\vskip -0.1in
\end{table}

We evaluated PAQclass using randomized 80-20 train-test splits. The 80-20 split seems to be the most common evaluation protocol used on this dataset. No document preprocessing was performed. The results are shown in Table~\ref{news}.

\begin{table*}[t]
\caption{Comparative results on the 20news dataset. Our results are in boldface.}
\label{news-comparison}
\vskip 0.15in
\begin{center}
\begin{footnotesize}
\begin{sc}
\addtolength{\tabcolsep}{-4pt}
{\begin{tabular}{lcccr}
\hline
\abovespace
\belowspace
Methodology & Protocol & Percent correct \\
\hline
\abovespace

extended version of Naive Bayes & 80-20 train-test split & $86.2$ \\
\belowspace
\citep{Rennie03} \\

SVM + error correcting output coding & 80-20 train-test split & $87.5$ \\
\belowspace
\citep{Rennie01} \\

language modeling & 80-20 train-test split & $89.23$ \\
\belowspace
\citep{Peng04} \\

AMDL using RAR compression & 80-20 train-test split & $90.5$ \\
\belowspace
\citep{Marton05} \\

multiclass SVM + linear kernel & 70-30 train-test split & $91.96$ \\
\belowspace
\citep{Weinberger09} \\

\belowspace
{\bf PAQclass} & {\bf 80-20 train-test split} & {\bf 92.35} \\

multinomial Naive Bayes + TFIDF & 80-20 train-test split & $93.65$ \\
\belowspace
\citep{Kibriya05} \\

\hline
\end{tabular}}
\end{sc}
\end{footnotesize}
\end{center}
\vskip -0.1in

\end{table*}

Our result of 92.3526\% correct is competitive with the best results published for this dataset. Table~\ref{news-comparison} shows some comparative results. It should be noted that there are several versions of the 20news dataset and many publications use different evaluation protocols. Some of these published results can not be directly compared. For example, Zhang \& Oles \citep{Zhang01} report a figure of 94.8\% correct on a version of the dataset in which the ``Newsgroup:'' headers were not removed from messages. The four best results (including PAQclass) in Table~\ref{news-comparison} \citep{Marton05,Weinberger09,Kibriya05} all seem to use the same version of 20news. PAQclass outperforms classification using the RAR compression algorithm \citep{Marton05} on this dataset.

\subsubsection{Shape Recognition}
\label{chicken-sec}

Shape recognition is the problem of assigning images to categories based on the shape or contour of an object within the image. We evaluated PAQclass on the chicken dataset of \citep{Andreu97}, available from \emph{http://algoval.essex.ac.uk/data/sequence/chicken}. This dataset contains 446 binary images of chicken parts in five categories (see Figure~\ref{chicken-classes}). Example images from this dataset are shown in Figure~\ref{chicken-figure}. The chicken pieces in the images are not set to a standard orientation. The images are square and vary in resolution from 556$\times$556 to 874$\times$874.

\begin{figure}[ht]
\vskip 0.1in
\begin{center}
\centerline{\includegraphics[width=3.5in]{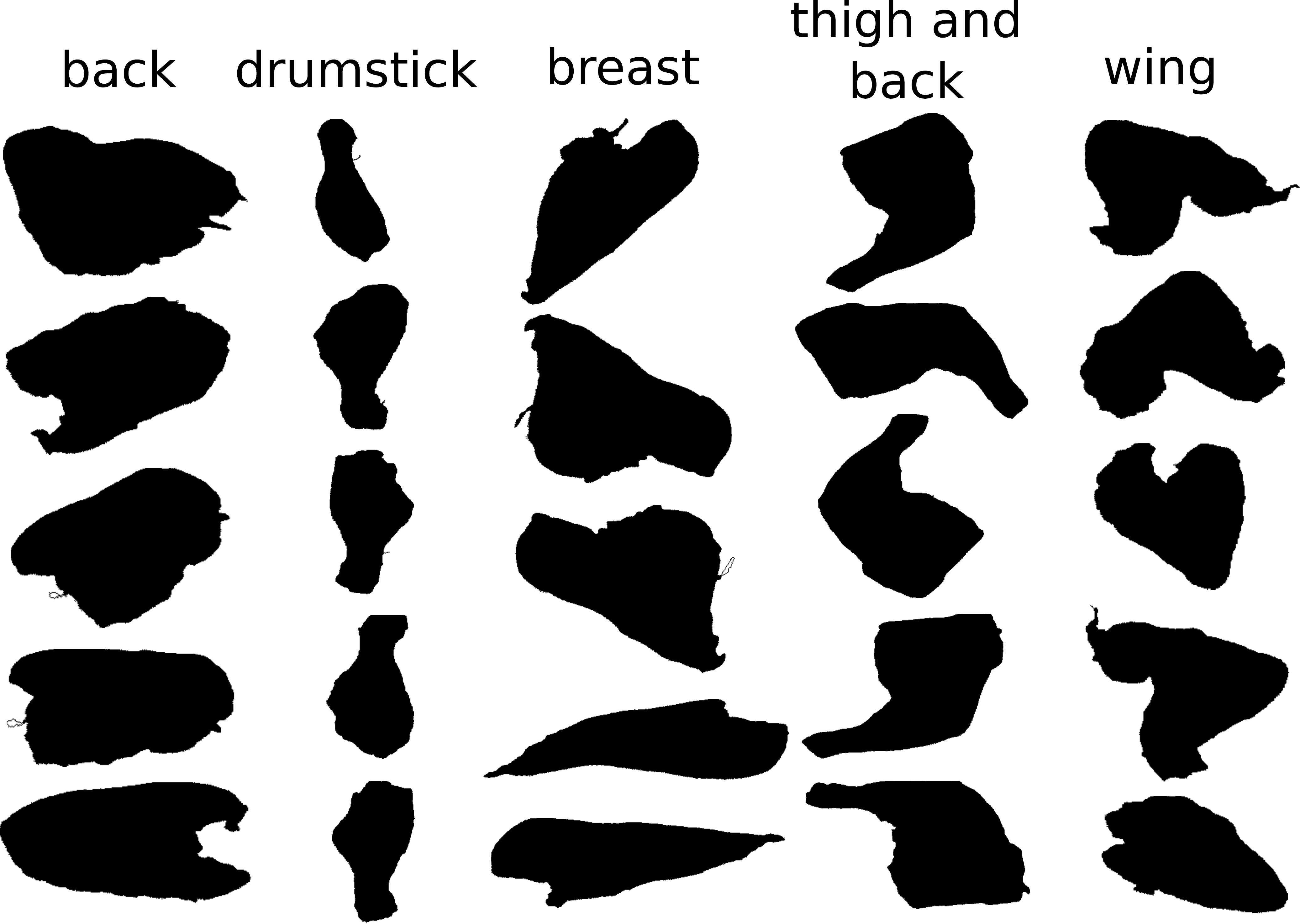}}
\caption{Five example images from each class of the chicken dataset. The images have not been rotated.}
\label{chicken-figure}
\end{center}
\vskip -0.2in
\end{figure}

\begin{table}[t]
\caption{Number of images in each class of the chicken dataset.}
\label{chicken-classes}
\vskip 0.15in
\begin{center}
\begin{small}
\begin{sc}
\begin{tabular}{lcccr}
\hline
\abovespace\belowspace
Class & Count \\
\hline
\abovespace
back & 76 \\
breast & 96 \\
drumstick & 96 \\
thigh and back & 61 \\
wing & 117 \\
\belowspace
{\bf Total} & {\bf 446} \\
\hline
\end{tabular}
\end{sc}
\end{small}
\end{center}
\vskip -0.1in
\end{table}

As discussed in Section~\ref{images}, compressing images poses a significantly different problem compared to compressing text. There does not seem to be a large body of research on using compression-based methods for image classification (in comparison to text categorization). This may be due to the fact that compression-based methods
tend to be slow and may be infeasible for the large datasets often used for object recognition tasks.

Lossy compression algorithms can be used for performing compression-based classification. There are several options for creating one-dimensional lossy representations of images. For example, Watanabe et~al \citep{Watanabe02} demonstrate a method of converting images to text. They show that their system is effective for image classification tasks. Wei et~al \citep{Wei08} describe a method of converting shape contours into time series data. They use this representation to achieve successful classification results on the chicken dataset. Based on these results, we decided to combine this representation with PAQ8 classification.
\begin{figure}[h!]
\vskip 0.1in
\begin{center}
\centerline{\includegraphics[width=2.4in]{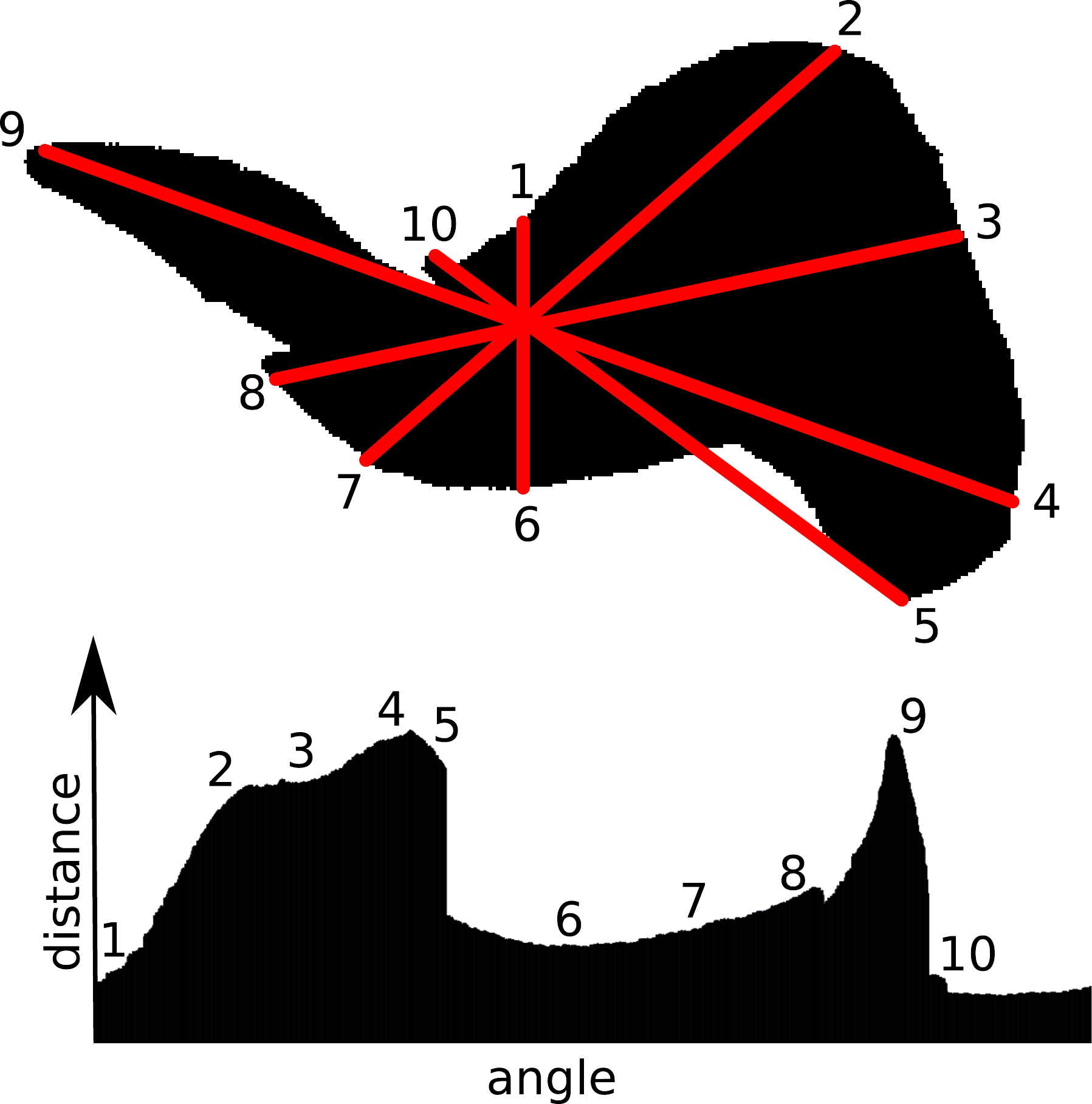}}
\caption{An example of converting a shape into one-dimensional time series data. The original image is shown on top (part of the ``wing'' class of the chicken dataset) and the time series data shown on the bottom. Points along the contour have been labeled and the corresponding points on the time series are shown.}
\label{time-figure}
\end{center}
\vskip -0.2in
\end{figure}
\begin{table}[h!]
\caption{Leave-one-out classification results on the chicken dataset with different settings of the ``number of measurements'' parameter. There are a total of 446 images. The row with the best classification results is in boldface.}
\label{chicken}
\vskip 0.15in
\begin{center}
\begin{small}
\begin{sc}
{\begin{tabular}{lcccr}
\hline
\abovespace
Number of & Correct & Percent \\
\belowspace
measurements & classifications & correct \\
\hline
\abovespace
1 & 162 & 36.3229 \\
5 & 271 & 60.7623 \\
10 & 328 & 73.5426 \\
30 & 365 & 81.8386 \\
35 & 380 & 85.2018 \\
38 & 365 & 81.8386 \\
39 & 363 & 81.3901 \\
{\bf 40} & {\bf 389} & {\bf 87.2197} \\
41 & 367 & 82.287 \\
42 & 367 & 82.287 \\
45 & 359 & 80.4933 \\
50 & 352 & 78.9238 \\
100 & 358 & 80.2691 \\
200 & 348 & 78.0269 \\
\belowspace
300 & 339 & 76.009 \\
\hline
\end{tabular}}
\end{sc}
\end{small}
\end{center}
\vskip -0.1in
\end{table}
\begin{table}[h!]
\renewcommand{\arraystretch}{1.3}
\caption{Confusion matrix for chicken dataset with the ``number of measurements'' parameter set to 40. C1=back, C2=breast, C3=drumstick, C4=thigh and back, C5=wing.}
\label{conf}
\begin{center}
{ \begin{tabular}{|c|c|c|c|c|c|c|c|c|}
\cline{3-7}
\multicolumn{2}{c}{}\vline &\multicolumn{5}{c}{Predicted}\vline\\
\cline{3-7}
\multicolumn{2}{c}{}\vline &C1&C2&C3&C4&C5\\
\hline
&C1&{\bf 55}&10&2&2&7\\
\cline{2-7}
&C2&0&{\bf 93}&0&3&0\\
\cline{2-7}
Actual&C3&0&5&{\bf 84}&0&7\\
\cline{2-7}
&C4&0&8&0&{\bf 48}&5\\
\cline{2-7}
&C5&3&1&3&1&{\bf 109}\\
\hline
\end{tabular}}
\end{center}
\end{table} 
\begin{table*}[h!]
\caption{Comparative results on the chicken dataset. Our results are in boldface.}
\label{comparison}
\vskip 0.15in
\begin{center}
\begin{footnotesize}
\begin{sc}
\addtolength{\tabcolsep}{-4pt}
{ \begin{tabular}{lcccr}
\hline
\abovespace
\belowspace
Methodology & Protocol & Percent correct \\
\hline
\abovespace
1-NN + Levenshtein edit distance & leave-one-out & $\approx 67$ \\
\belowspace
\citep{Mollineda02} \\
1-NN + HMM-based distance & leave-one-out & $73.77$ \\
\belowspace
\citep{Bicego08} \\
1-NN + mBm-based features & leave-one-out & $76.5$ \\
\belowspace
\citep{Bicego08} \\
1-NN + approximated cyclic distance & leave-one-out & $\approx 78$ \\
\belowspace
\citep{Mollineda02} \\

1-NN + convert to time series & leave-one-out & $80.04$ \\
\belowspace
\citep{Wei08} \\
SVM + HMM-based entropic features & leave-one-out & $81.21$ \\
\belowspace
\citep{Perina09} \\
SVM + HMM-based nonlinear kernel & 50-50 train-test split & $85.52$ \\
\belowspace
\citep{Carli09} \\
SVM + HMM-based Fisher kernel & 50-50 train-test split & $85.8$ \\
\belowspace
\citep{Bicego09} \\

\belowspace
{\bf PAQclass + convert to time series} & {\bf leave-one-out} & {\bf 87.22} \\

\hline
\end{tabular}
}
\end{sc}
\end{footnotesize}
\end{center}
\vskip -0.1in
\end{table*}

Figure~\ref{time-figure} demonstrates how we convert images in the chicken dataset into one-dimensional time series data. The first step is to calculate the centroid of the shape. We project a ray from the centroid point and measure the distance to the edge of the shape. If the ray intersects the shape edge at multiple points, we take the furthest intersection (as seen at point 5 in Figure~\ref{time-figure}). We rotate the ray around the entire shape and take measurements at uniform intervals. The number of measurements taken along the shape contour is a tunable parameter. Once the Euclidean distance is measured for a particular angle of the ray, it is converted into a single byte by rounding the result of the following formula: $(100*distance/width)$. $distance$ is the Euclidean distance measurement and $width$ is the width of the image. PAQclass is then run on this binary data.

The classification results for different settings of the ``number of measurements'' parameter are shown in Table~\ref{chicken}. We used leave-one-out cross-validation since this seems to be the most common evaluation protocol used on this dataset. The ``number of measurements'' parameter had a surprisingly large effect on classification accuracy. Adjusting the parameter by a single measurement from the best value (40) resulted in $\approx5$ to $6\%$ loss in accuracy. Another unfortunate property of adjusting this parameter is that the classification accuracy is not a convex function (as seen at the parameter value 35). This means finding the optimal value of the parameter would require an exhaustive search. Due to time constraints, we did not perform an exhaustive search (only the experiments in Table~\ref{chicken} were performed). Table~\ref{conf} shows a confusion matrix at the best parameter setting.

Our result of 87.2197\% correct classifications is among the best results published for this dataset. Table~\ref{comparison} shows some comparative results.

The classification procedure we used is not rotationally invariant. Since the chicken pieces in the dataset can be in arbitrary orientations, this could lead to a decrease in classification accuracy. Wei et~al \citep{Wei08} use the same one dimensional image representation as we use, but they use a rotationally invariant classification procedure. They use a 1-nearest-neighbor classifier combined with a Euclidean distance metric. When comparing the distance between two images, they try all possible rotations and use the angle which results in the lowest Euclidean distance between the time series representations. This same procedure of trying all possible orientations could be used to make the PAQclass classification procedure rotationally invariant (although it would increase the algorithm's running time). Alternatively, we could use a single rotationally invariant representation such as always setting the smallest sampled edge distance to be at $angle=0$. The effect of rotational invariance on classification accuracy would make interesting future work.
\subsection{Lossy Compression}
\begin{figure}[h!]
\vskip 0.1in
\begin{center}
\centerline{\includegraphics[width=0.6\textwidth]{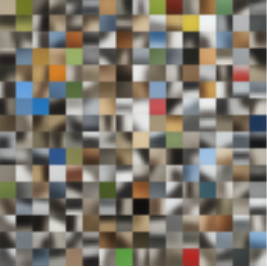}}
\caption{256 6$\times$6 image filters trained using k-means on the CIFAR-10 dataset.}
\label{fig:filters}
\end{center}
\vskip -0.2in
\end{figure}
\begin{figure}[h!]
\vskip 0.1in
\begin{center}
\centerline{\includegraphics[width=0.7\textwidth]{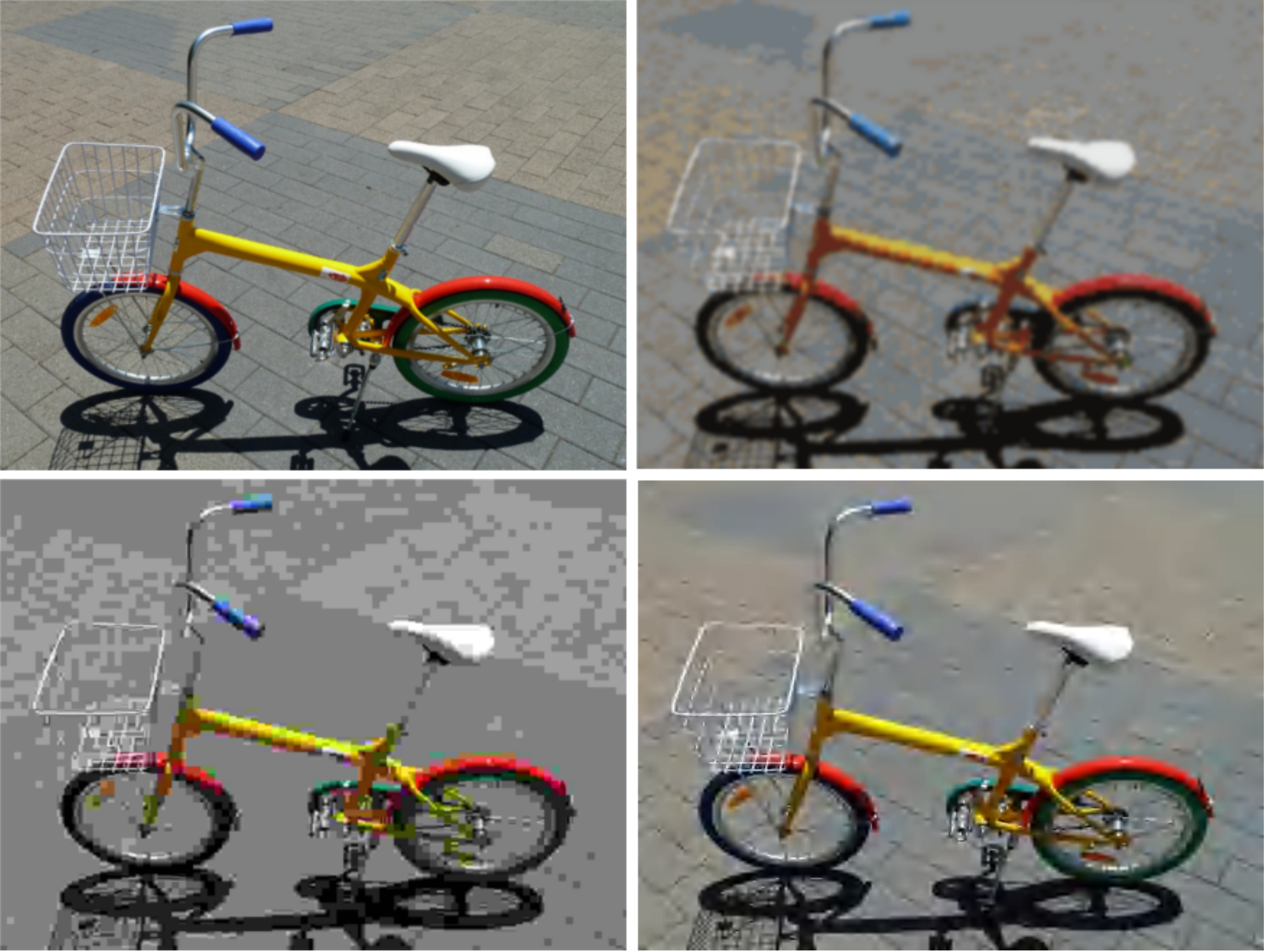}}
\caption{top-left image: original (700$\times$525 pixels), top-right image: our compression method (4083 bytes), bottom left: JPEG (16783 bytes), bottom-right: JPEG2000 (4097 bytes)}
\label{fig:lossy1}
\end{center}
\vskip -0.2in
\end{figure}
\begin{figure}[h!]
\vskip 0.1in
\begin{center}
\centerline{\includegraphics[width=0.5\textwidth]{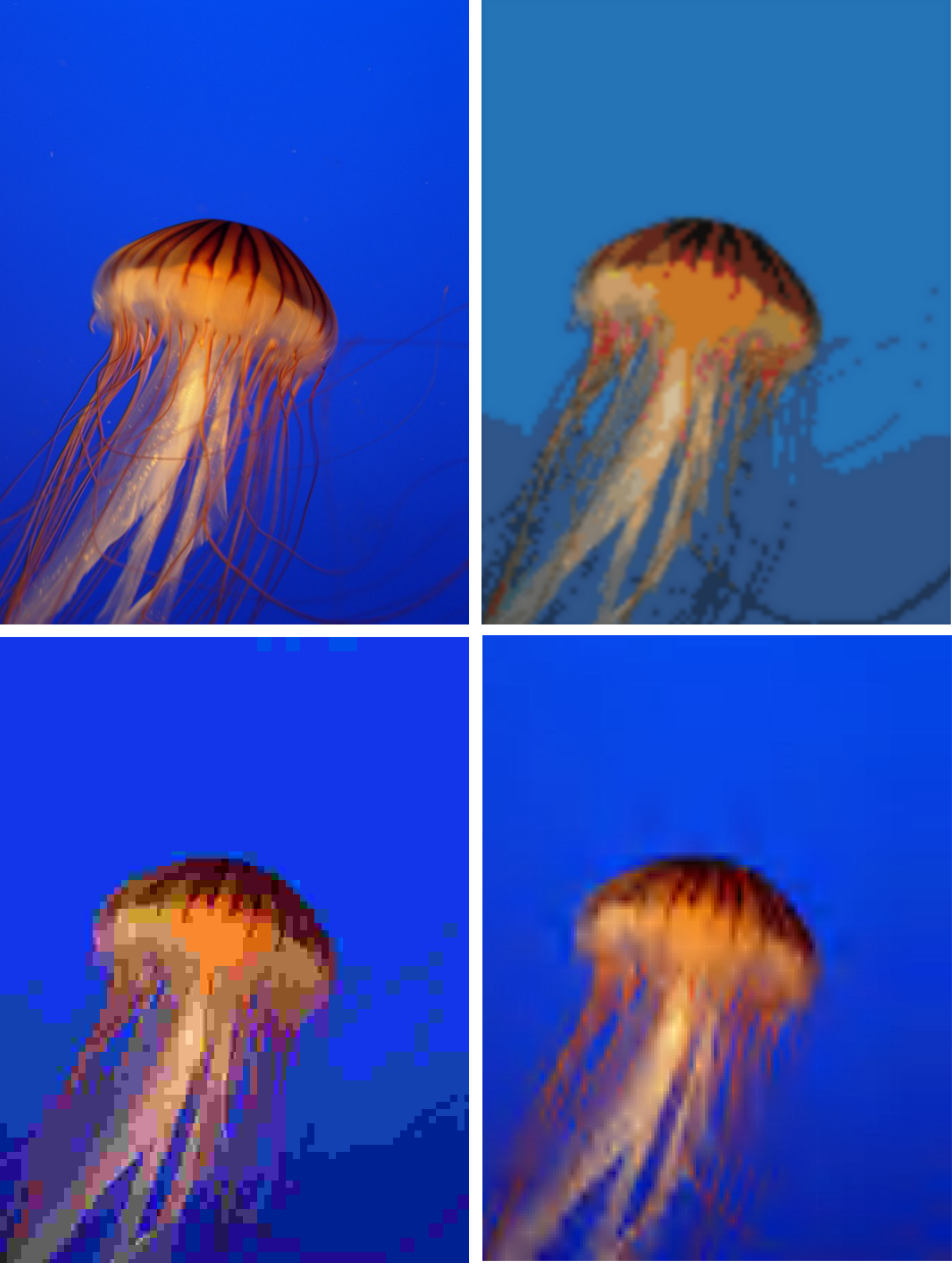}}
\caption{top-left image: original (525$\times$700 pixels), top-right image: our compression method (1493 bytes), bottom left: JPEG (5995 bytes), bottom-right: JPEG2000 (1585 bytes)}
\label{fig:lossy2}
\end{center}
\vskip -0.2in
\end{figure}
\begin{figure}[h!]
\vskip 0.1in
\begin{center}
\centerline{\includegraphics[width=0.7\textwidth]{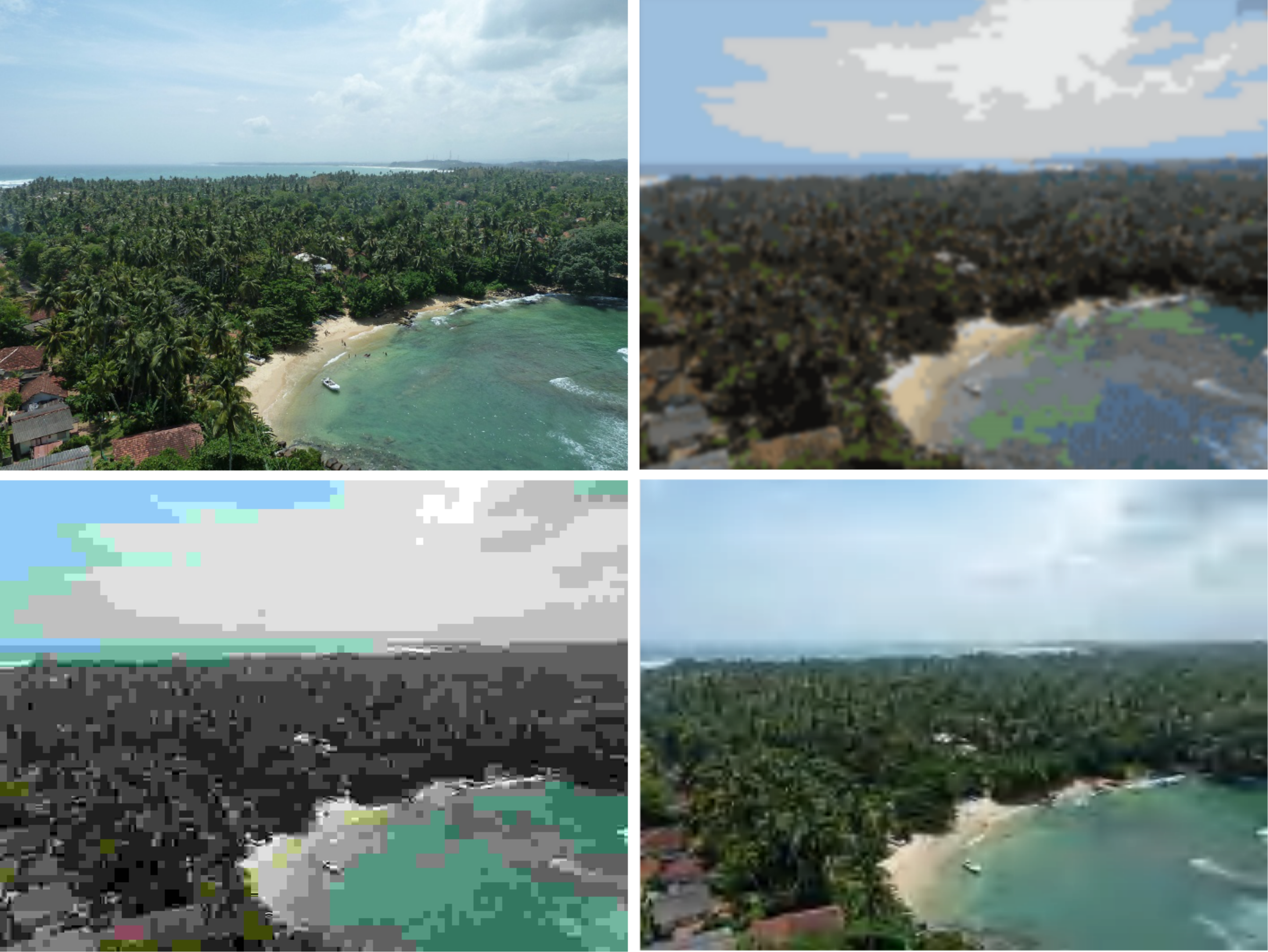}}
\caption{top-left image: original (700$\times$525 pixels), top-right image: our compression method (3239 bytes), bottom left: JPEG (16077 bytes), bottom-right: JPEG2000 (2948 bytes)}
\label{fig:lossy3}
\end{center}
\vskip -0.2in
\end{figure}
\begin{figure}[h!]
\vskip 0.1in
\begin{center}
\centerline{\includegraphics[width=0.7\textwidth]{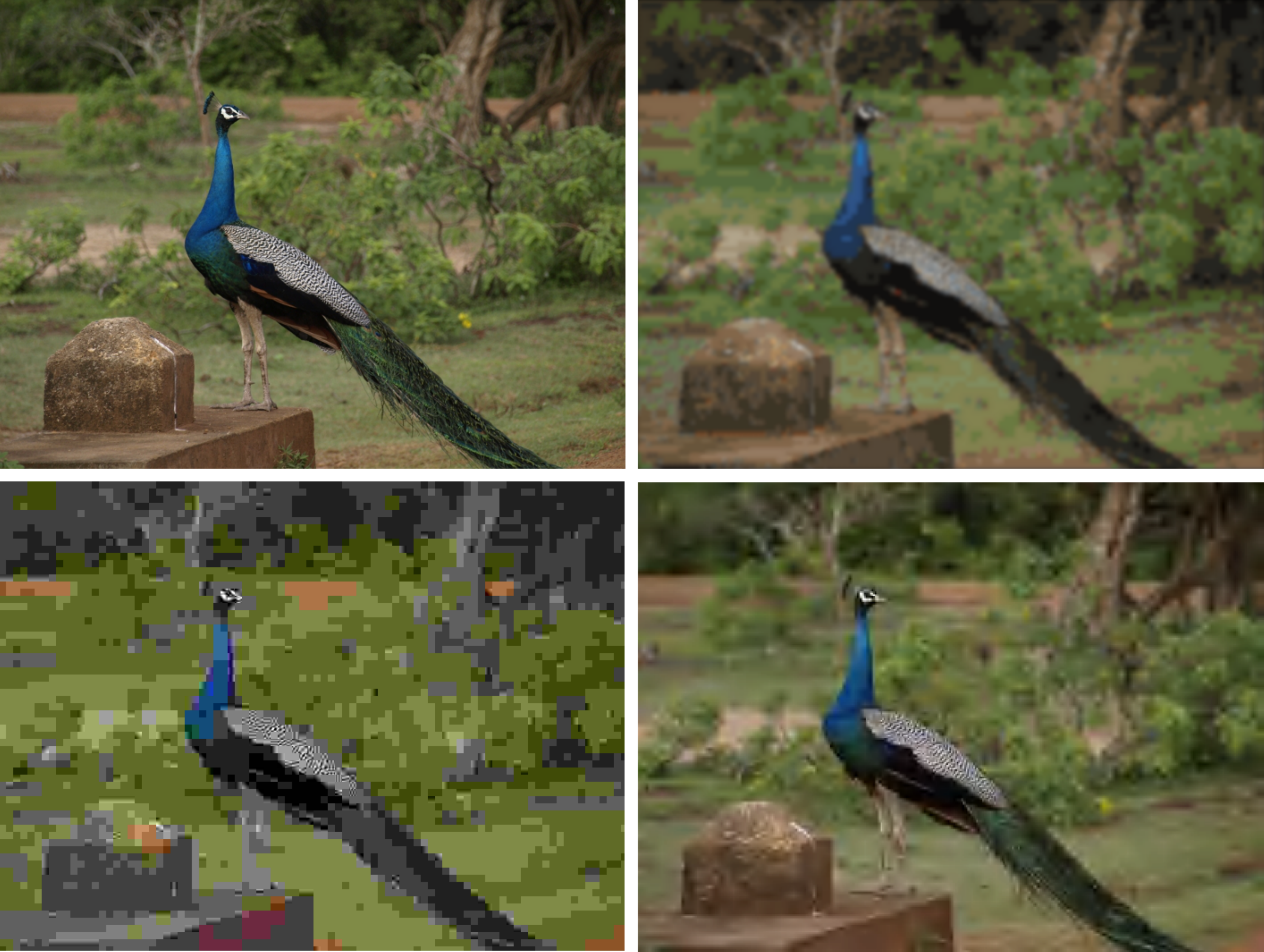}}
\caption{top-left image: original (700$\times$525 pixels), top-right image: our compression method (3970 bytes), bottom left: JPEG (6335 bytes), bottom-right: JPEG2000 (3863 bytes)}
\label{fig:lossy4}
\end{center}
\vskip -0.2in
\end{figure}
PAQ8 can also be used for performing lossy compression. Any lossy representation can potentially be passed through PAQ8 to achieve additional compression. For example, {\tt paq8l} can losslessly compress JPEG images by about 20\% to 30\%. {\tt paq8l} contains a model specifically designed for JPEG images. It essentially undoes the lossless compression steps performed by JPEG (keeping the lossy representation) and then performs lossless compression more efficiently.
To create a lossy image compression algorithm, we first created a set of filters based on a method described by Coates et~al \citep{Coates2010}. We used the k-means algorithm to learn a set of 256 6$\times$6 filters on the CIFAR-10 image dataset \citep{Alex09}. The filters were trained using 400,000 randomly selected images patches. The filters are shown in Figure~\ref{fig:filters}.

In order to create a lossy image representation, we calculated the closest filter match to each image patch in the original image. These filter selections were encoded by performing a raster scan through the image and using one byte per patch to store the filter index. These filter selections were then losslessly compressed using {\tt paq8l}. Some example images compressed using this method are shown in Figures \ref{fig:lossy1}, \ref{fig:lossy2}, \ref{fig:lossy3}, and \ref{fig:lossy4}. At the maximum JPEG compression rate, the JPEG images were still larger than the images created using our method. Even at a larger file size the JPEG images appeared to be of lower visual quality compared to the images compressed using our method. We also compared against the more advanced lossy compression algorithm JPEG2000. JPEG2000 has been designed to exploit limitations of human visual perception: the eye is less sensitive to color variation at high spatial frequencies and it has different degrees of sensitivity to brightness variation depending on spatial frequency \citep{Mahoney11}. Our method was not designed to exploit these limitations (we leave this as future work). It simply uses the filters learned from data. Based on the set of test images, JPEG2000 appears to outperform our method in terms of visual quality (at the same compression rate).


\section{Conclusion}
\label{ch:Conclusion}

We hope this technical exposition of PAQ8 will make the method more accessible and stir up new research in the area of temporal pattern learning and prediction. Casting the weight updates in a statistical setting already enabled us to make modest improvements to the technique. We tried several other techniques from the fields of stochastic approximation and nonlinear filtering, including the unscented Kalman filter, but did not observe significant improvements over the {EKF} implementation. One promising technique from the field of nonlinear filtering we have not yet implemented is Rao-Blackwellized particle filtering for online logistic regression \citep{Andrieu01}. We leave this for future work.

The way in which PAQ8 \emph{adaptively} combines predictions from multiple models using context matching is different from what is typically done with mixtures of experts and ensemble methods such as boosting and random forests. A statistical perspective on this, which allows for a generalization of the technique, should be the focus of future efforts. Bridging the gap between the online learning framework \citep{Cesa-Bianchi2006} and PAQ8 is a potentially fruitful research direction. Recent developments in RNNs seem to be synergistic with PAQ8, but this still requires methodical exploration. Of particular relevance is the adoption of PAQ8's deterministic gating architecture so as to reduce the enormous computational cost of training RNNs. This should be done in conjunction with a move toward adaptive prediction.

On the application front, we found it remarkable that a single algorithm could be used to tackle such a broad range of tasks. In fact, there are many other tasks that could have been tackled, including clustering, compression-based distance metrics, anomaly detection, speech recognition, and interactive interfaces. It is equally remarkable how the method achieves comparable results to state-of-the-art in text classification and image compression.

There are challenges in deploying PAQ beyond this point. The first challenge is that the models are non-parameteric and hence require enormous storing capacity. A better memory architecture, with some forgetting, is needed. The second challenge is the fact that PAQ applies only to univariate sequences. A computationally efficient extension to multiple sequences does not seem trivial. In this sense, RNNs have an advantage over PAQ, PPM and stochastic memoizers.

\section*{Acknowledgements}

We would like to thank Matt Mahoney, who enthusiastically helped us understand important details of PAQ and provided us with many insights into predictive data compression. We would also like to thank Ben Marlin and Ilya Sutskever for discussions that helped improve this manuscript.


\appendix

\section{PAQ8 Demonstrations}
\label{ch:a}

Two of the applications are available at: \emph{http://cs.ubc.ca/$\sim$knoll/PAQ8-demos.zip}. The first demonstration is text prediction and the second demonstration is a rock-paper-scissors AI. Instructions are provided on how to compile and run the programs in Linux.



\bibliographystyle{abbrvnat}
\bibliography{biblio}

\end{document}